\documentclass[final,5p,times,twocolumn]{elsarticle}

\usepackage{amsmath,amssymb,amsfonts}
\usepackage{xcolor}
\usepackage{enumitem}
\usepackage{graphicx}
\usepackage{booktabs}
\usepackage{algorithm}
\usepackage{algpseudocode}

\usepackage{microtype}
\usepackage{float}
\usepackage{hyperref}

\usepackage{lipsum}  
\usepackage{comment}
\usepackage{array}
\usepackage{longtable}

\journal{jour}

\begin{document}

\begin{frontmatter}



\title{Flexible Manufacturing Systems Intralogistics: Dynamic Optimization of AGVs and Tool Sharing Using Coloured-Timed Petri Nets and Actor-Critic RL with Actions Masking}


\author[inst1]{Sofiene Lassoued}
\author[inst3]{Laxmikant Shrikant Baheti}
\author[inst2]{Nathalie Weiß-Borkowski}
\author[inst3]{Stefan Lier}
\author[inst1]{Andreas Schwung}

\affiliation[inst1]{organization={South Westphalia University of Applied Sciences}, 
                    Department {Automation Technology and Learning Systems}, 
                    addressline={Lübecker Ring 2}, 
                    city={Soest}, 
                    postcode={59494}, 
                    state={North Rhine-Westphalia}, 
                    country={Germany}}

\affiliation[inst2]{organization={South Westphalia University of Applied Sciences}, 
                    department={Materials Science Laboratory and 3D Printing Center}, 
                    addressline={Lübecker Ring 2}, 
                    city={Soest}, 
                    postcode={59494}, 
                    state={North Rhine-Westphalia}, 
                    country={Germany}}                    

\affiliation[inst3]{organization={South Westphalia University of Applied Sciences}, 
                    Department {Logistik und Supply Chain Management}, 
                    addressline={Lindenstr. 53}, 
                    city={Meschede},   
                    postcode={59872}, 
                    state={North Rhine-Westphalia}, 
                    country={Germany}}

\begin{abstract}

Flexible Manufacturing Systems (FMS) are pivotal in optimizing production processes in today's rapidly evolving manufacturing landscape. This paper advances the traditional job shop scheduling problem by incorporating additional complexities through the simultaneous integration of automated guided vehicles (AGVs) and tool-sharing systems. We propose a novel approach that combines Colored-Timed Petri Nets (CTPNs) with actor-critic model-based reinforcement learning (MBRL), effectively addressing the multifaceted challenges associated with FMS. CTPNs provide a formal modeling structure and dynamic action masking, significantly reducing the action search space, while MBRL ensures adaptability to changing environments through the learned policy. Leveraging the advantages of MBRL, we incorporate a lookahead strategy for optimal positioning of AGVs, improving operational efficiency. Our approach was evaluated on small-sized public benchmarks and a newly developed large-scale benchmark inspired by the Taillard benchmark. The results show that our approach matches traditional methods on smaller instances and outperforms them on larger ones in terms of makespan while achieving a tenfold reduction in computation time. To ensure reproducibility, we propose a gym-compatible environment and an instance generator. Additionally, an ablation study evaluates the contribution of each framework component to its overall performance.

\end{abstract}


\end{frontmatter}

\section{Introduction}

In today's competitive manufacturing landscape, flexibility and responsiveness are crucial. Traditional manufacturing, focusing on high-volume, low-variety output for cost efficiency, often struggles to adapt to changing market demands. This highlights the need for more adaptable manufacturing approaches \cite{ElMaraghy.2005}. Flexible Manufacturing Systems (FMS) provide the necessary flexibility and efficiency, enabling manufacturers to produce various products swiftly and cost-effectively with minimal manual intervention. A key aspect of FMS is the ability to adjust machine schedules dynamically, reroute jobs, and reallocate resources in response to changes in production requirements or disruptions. This design promotes rapid responsiveness to change, making FMS vital for addressing the complexities and uncertainties of modern manufacturing environments \cite{Chen.1991}.

The connection between FMS and intralogistics is fundamental to achieving efficient internal workflows. Intralogistics focuses on the internal movement of materials, information, and products within a facility \cite{Fottner.2021}. Its integration within FMS is crucial for optimizing workflows and increasing production efficiency \cite{Fernandes.2019}. To explore this relationship further, the authors of \cite{MartinezBarbera.2010} utilized Discrete Event Simulation (DES) to analyze the impact of AGVs on factory and transport efficiency within FMS. The findings demonstrate that AGVs' parameters significantly influence FMS performance, revealing a correlation between transport efficiency and machine utilization. Similar to AGVs, the authors in \cite{kashyap1996} conducted simulations to investigate factors related to tool sharing for enhancing system efficiency. They concluded that, in addition to tool duplication, the rules for selecting tools significantly impact both the makespan and the utilization of tool transporters.

However, integrating AGVs and tool-sharing within traditional JSSPs presents challenges. The added complexity of an increased action space complicates decisions on AGVs and tool transporter assignments, particularly in large setups. While JSSP operation sequences are predefined, AGVs and tool transporter routing rely on these sequences, making scheduling and routing interdependent and thus more complex.

Despite advancements, existing methods often fall short, particularly in integrating scheduling with intralogistics and adapting to the dynamic nature of FMS. Current optimization approaches face limitations, including those based on exact solutions, heuristics, and metaheuristics. For instance, exact solutions are impractical for large instances due to computational constraints, and heuristics often require domain knowledge, resulting in suboptimal solutions \cite{vanEkeris.2021}. Metaheuristics \cite{Engin.2018, Yu.2018,Umam.2022} are more general but encounter difficulties with hyperparameter tuning and adapting to dynamic environments.

Reinforcement learning (RL), particularly deep reinforcement learning (DRL), has emerged as a promising approach for handling complex, dynamic systems. RL methods like those developed in \cite{Chen.2022,Hameed.2023,Luo.2020} allow for exploration combined with knowledge retention, making them suitable for FMS scheduling tasks. Using Petri Nets to model these environments provides a robust framework, capturing concurrency and resource-sharing constraints \cite{Tuncel.2007}. When enhanced with methods like Graph Neural Networks, the modeling capabilities expand further \cite{Hu.2020}.

Our approach innovates by incorporating intralogistics elements such as AGVs and tool sharing, leveraging Colored-Timed Petri Nets with RL to manage the expanded action space using action masking. This integration enables us to test large FMS instances and address existing gaps in the literature on large-scale optimization. Our approach, built on an actor-critic policy that supports knowledge retention and adaptability, is well-suited for dynamic environments. Furthermore, the inherent modularity of the Petri net model ensures flexibility and robustness, making it particularly effective for FMS applications.

\bigskip
The contribution of this study can be summarized as follows:
\begin{enumerate}
   \item We propose an innovative methodology to address the challenges of FMS with AGVs and Tool sharing. We propose a modular approach by integrating colored-Timed Petri Net modeling with actor-critic RL techniques. This approach allows for the plug-in and out of AGVs and tool management, combining the strengths of formal process representation with adaptive decision-making.  \\

   \item Leveraging the strengths of model-based RL in conjunction with the Petri Net model of FMS, we introduce the Lookahead technique. By simulating future steps in a twin environment, this approach ensures that AGVs are optimally positioned in advance and ready to transport the pieces efficiently, and reduces the makespan.   \\
    
  \item We evaluate the dynamic adaptability and generalization capacity of our approach. The generalization enables a single agent to handle instances of various sizes. The dynamic capability allows the system to handle new orders during deployment, simulating real-world scenarios such as late order arrivals. \\

    \item We introduce a new benchmark, inspired by the Taillard benchmark, to address the gap in the literature for large-scale instances involving AGVs and tool sharing. To support this, we provide a gym-compatible environment and an instance generator, ensuring reproducibility and facilitating better benchmarking.\\
    
  \item We conduct a comparative study on public benchmarks, comparing our results to heuristics and metaheuristics. Furthermore, we delve into ablation studies to analyze the contribution of individual elements of the framework.\\

\end{enumerate}

This paper is structured as follows: Section \ref{section: Literature Review} provides a comprehensive review of related works, establishing the foundation for our study. Section \ref{section: Preliminaries} introduces fundamental concepts, including Colored-Timed Petri Nets and RL. Section \ref{section: Problem and Modelling} outlines our research problem, the key assumptions, and constraints we are operating under. Section \ref{section: Methodology and Implementation} details our methodology, beginning with the description of the FSM model using Colored-Timed Petri nets, followed by the integration of the Maskable PPO agent and the lookahead approach. Section \ref{section: Results and Discussion} presents a new large-scale benchmark and compares our approach with competing algorithms, analyzing the training behavior and providing an in-depth examination of the results. Additionally, we evaluate our approach's generalization and dynamic capabilities along with an ablation study to assess the contributions of various components within our method. Finally, Section \ref{section: Conclusion} summarizes the key insights gained from this study and suggests promising avenues for future research.

\section{Literature Review}
\label{section: Literature Review}

This paper addresses the challenges of FMS, including machine scheduling, AGVs routing, and tools management, by combining Petri Nets for formal modeling with Model-Based RL for dynamic decision-making. The literature review covers Petri Nets in production scheduling, heuristics and metaheuristics in scheduling optimization, and advancements in RL in scheduling applications. Then, it examines the literature on AGVs integration and tool management in FMS, concluding with an analysis of research gaps and challenges.

Petri Nets are a powerful mathematical tool for modeling concurrent systems, efficiently representing states, precedence relationships, and key aspects like deadlocks, conflicts, and synchronization in production systems \cite{Tuncel.2007}. Over time, extensions to Petri Nets have expanded their functionality to address more complex scenarios. For instance, Stochastic Petri Nets facilitate the representation and analysis of probabilistic elements \cite{bause2002,Hatono.1991}. Similarly, Coloured Petri Nets introduce diverse token types within a single net, enabling more compact and expressive networks \cite{Jensen.1989}. Additionally, Timed Petri Nets incorporate time-related features, allowing for modeling delays and temporal relationships, which are essential for accurately simulating scheduling problems \cite{Shih.1991}. Petri Nets serve three key functions in industrial applications. First, they predict and manage system states, including deadlock prevention, through reachability graphs \cite{Mejia.2018}. Second, their robust mathematical foundation supports both qualitative and quantitative analyses, providing valuable insights into system dynamics \cite{zhou1999}. Finally, Petri Nets act as important simulation tools, working alongside optimization algorithms to enhance system efficiency and performance \cite{Doo.1994}.

Optimization algorithms used in scheduling can be categorized into exact and approximate solutions. Exact methods, as discussed in \cite{Wagner.1959,Brucker.1994,jamili.2016}, guarantee optimal solutions but are often computationally intensive and limited by a restricted action space. In contrast, approximate solutions, which include heuristics, metaheuristics, and iterative approaches like RL, often yield suboptimal solutions but with significantly reduced computational time. Heuristics, as implemented in \cite{naderi2014}, are widely used in scheduling due to their simplicity and lower computational cost compared to exact methods. However, their efficiency is often compromised by the generation of sub-optimal solutions and a reliance on domain-specific knowledge, limiting their general applicability. Metaheuristics, such as Genetic Algorithms (GA) and Ant Colony Optimization (ACO), draw inspiration from natural processes and provide high-level strategies for solving JSSPs \cite{Yu.2018,Engin.2018}. Tabu Search, another well-known metaheuristic introduced by Glover \cite{Glover.1989}, helps escape local optima by using memory to avoid revisiting previous solutions.


Among approximate optimization approaches, RL, particularly with deep reinforcement learning (DRL) techniques that combine RL with deep neural networks, has gained significant attention in solving scheduling problems \cite{silver2018}. RL has proven effective in dynamic environments, with 85\% of implementations showing notable improvements in JSSP scheduling performance \cite{Panzer.2022}. Approaches such as Deep Q-Networks (DQN) \cite{Luo.2020} and Duelling Double DQN with Prioritized Replay (DDDQNPR) \cite{Han.2020} excel in dynamic environments, particularly with uncertain processing times. Multi-agent actor-critic methods, including Deep Deterministic Policy Gradient (DDPG), have also been successfully applied to JSSPs \cite{Liu.2020}. Hierarchical RL and graph-based methods, such as multi-pointer graph networks \cite{Lei.2024} and multi-proximal policy optimization \cite{Lei.2022}, have shown effectiveness in decomposing complex problems and improving scheduling efficiency. Graph Neural Networks (GNN) \cite{Zhang.2020,Hameed.2023} enhance generalization capabilities on large-scale instances, while the Gated-Attention model \cite{Gebreyesus.2023,Chen.2023} improves performance by focusing on relevant features. Additionally, transformer-based models that use node2vec and multi-head attention \cite{chen2022} have demonstrated promise for large-scale JSSP problems, offering long-range dependencies and parallel computing capabilities.

The survey conducted by Xiong et al. \cite{xiong2022} predicts that scheduling involving robots and AGVs will become a key area of research. Bilge and Uluyos \cite{Bilge.1995} expanded the scope of the JSSP by introducing material transport through AGVs. Their benchmarks, which integrated transport and job sequencing, have gained significant attention, especially with the rise of automated manufacturing systems. In their study \cite{heger2019reducing}, the authors aimed to reduce mean tardiness in FMS with AGVs by optimizing combinations of sequencing and routing rules. They continued their research in \cite{heger2019dynamic}, where they evaluated 27 combinations of rules, focusing on dynamic priority-based dispatching of AGVs. To address the Flexible Job Shop Scheduling Problem with Transportation Times and Many Robots (FJSPT–MR), the authors in \cite{nouri2016} employed hybrid metaheuristics based on a clustered holonic multi-agent model. They initially used a Neighborhood-based Genetic Algorithm (NGA) for global exploration of the search space, followed by a tabu search performed by cluster agents to refine results in processing areas. Finally, the authors in \cite{lacomme2016} proposed a method for solving a JSSP with transportation constraints, utilizing a master-slave approach. The master scheduling problem was modeled as a disjunctive graph, while the AGVs routing problem was treated as the slave problem.

Many research papers focus on scheduling problems using various optimization methods. However, as previously mentioned, only a few have addressed the added complexity of AGVs. Few have explored the simultaneous integration of Job Shop Scheduling Problems with Transport and Tools (JSSPT-T). This integration requires optimizing job operation sequences, tool transportation, and AGVs material transportation in a coordinated manner. Notably, the study by Reddy et al.\cite{Reddy.2021} utilized the Symbiotic Organism Search (SOS) algorithm to tackle this multifaceted challenge. By leveraging its symbiotic mechanisms, the algorithm efficiently navigates the solution space, minimizing the risk of premature convergence and improving its ability to identify global optima. As a result, it outperforms traditional metaheuristics regarding solution quality and computational efficiency.


Our literature review highlights three key gaps this paper seeks to address: the lack of integration in tackling the multifaceted challenges of FMS by optimizing the machine scheduling, AGVs routing, and tools management in a unified and cohesive framework, the need for large-scale benchmark and the corresponding scalable solutions, and the absence of knowledge retention mechanisms to address dynamic environments.

Current approaches often isolate critical components of FMS, such as machine scheduling, AGV routing, and tool sharing, treating them as standalone problems rather than addressing their interconnected nature. Furthermore, most studies are constrained to small-scale benchmarks like the Bilge benchmark, which fails to evaluate algorithm performances in complex, large-scale scenarios, underscoring the need for a larger-scale benchmark. Lastly, existing optimization methods heavily rely on exhaustive search techniques, which are computationally expensive and lack the ability to retain and utilize learned knowledge, resulting in inefficiencies when the environment changes. To overcome these challenges, we propose a cooperative framework that integrates Petri Nets for modeling with RL for decision-making. By leveraging action masking, this approach reduces the search space, enhances scalability, and adapts efficiently to dynamic environments, providing a comprehensive solution for solving large-scale JSSPs involving AGVs and tools.

\section{Background and Preliminaries }
\label{section: Preliminaries}

This section introduces the theories behind our approach's core elements: the RL framework and  Petri Nets. 

\subsection{Reinforcement Learning}
\label{section: RL Preliminaries}
Reinforcement learning is a branch of machine learning that trains agents to make sequential decisions within an environment to maximize cumulative rewards. RL operates within the Markov Decision Processes (MDPs) framework, defined by the tuple (\textit{S}, \textit{A}, \textit{P}, \textit{R},$\gamma$). \textit{S} represents the set of possible states; \textit{A} is the set of available actions, and \textit{P} is the state transition probability function $\textit{P}(s' |s, a)$, which describes the likelihood of transitioning from state $s$ to state $s'$ by taking action $a$. The reward function $\textit{R}(s, a, s')$ provides the immediate reward for moving from state $s$ to state $s'$ via action $a$. The discount factor $\gamma$ balances the importance of immediate versus future rewards in the expected return. The goal of RL algorithms is to learn an optimal policy $\pi^*$ that maximizes the expected cumulative discounted rewards. The reward function  $\mathit{G_t}$ is given by :

\begin{equation}
\label{eq:return}
G_t = \sum_{k=0}^{\infty} \gamma^{k} R_{t+k+1}.
\end{equation}

 The RL framework is built on the continuous interaction between the environment and the optimization agent. The agent's primary goal is to learn the optimal policy that achieves the highest possible reward over time \cite{Sutton.2018}. 

Learning the optimal policy in RL can be achieved using value-based, policy-based, or hybrid approaches. In value-based methods, the agent learns to predict the value of being in a specific state or taking a particular action in a given state, typically using the state-value function $\textit{V}(s)$ or the state-action-value function $\textit{Q}(s, a)$, also known as quality function. Based on these estimates, the agent uses a policy such as $\epsilon$-greedy to select actions, balancing exploring new actions with exploiting known rewards to find the optimal strategy. In contrast, policy-based methods directly optimize the policy by learning a mapping from states to actions without explicitly estimating value functions. Hybrid methods, like actor-critic, combine both approaches using a value function to guide policy optimization. Each method has strengths: value-based methods are efficient for discrete action spaces, while policy-based methods can handle continuous action spaces more effectively.

In this study, we employ a modified variant of the actor-critic algorithm known as Maskable Proximal Policy Optimization (MPPO). The rationale behind selecting MPPO is twofold. First, compared to the standard actor-critic approach, Proximal Policy Optimization (PPO) \cite{schulman2017proximal} introduces a clipping mechanism that constrains the policy update step, ensuring that the new policy does not deviate significantly from the old one. This promotes training stability, which is especially important in complex environments. Second, the maskable extension of PPO allows the integration of action masks, a Boolean vectors that indicate valid actions at each step. In our case, these masks are derived from the Petri Net guard functions, which encode domain constraints. This allows the RL agent to drastically reduce the action space by ignoring infeasible actions, a valuable feature for solving NP-hard combinatorial problems like the JSSP.

\subsection{Petri Nets}
As the RL introduction section \ref{section: RL Preliminaries} outlines, the RL framework consists of two main components: the agent and the environment. The environment can represent the real world or be a simulated model, enabling more controlled and efficient experimentation. This paper uses a Petri Net to model a flexible manufacturing system, positioning our approach as model-based RL. By leveraging the structured Petri Net model, we gain key advantages of MBRL, including improved sample efficiency, as fewer interactions with the system are needed to learn optimal policies; enhanced interpretability, with a clear representation of the system's dynamics such as concurrency and resource constraints; and the ability to conduct simulated experiments, which allow for controlled testing and experimentation without the need for costly real-world trials.

Petri Nets are formal models that describe and analyze dynamic systems characterized by concurrency and synchronization. They offer graphical and mathematical tools to represent system behavior, allowing for evaluating key properties such as reachability, liveness, and deadlock prevention \cite{Mejia.2018}. Formally, a Petri Net can be defined as the pair $(\mathcal{G}, \mu_0)$, where $\mathcal{G}$ is a directed bipartite graph made up of a finite set of nodes and edges, and $\mu_0$ represents the initial marking, which specifies the initial distribution of tokens across the places in the net. Tokens are abstract representations of system resources or job states that move between places, enabling the firing of transitions and indicating the system's dynamic state.

Unlike ordinary graphs, Petri Nets have a unique structure in which nodes are split into two disjoint groups: places $\mathcal{P}$ and transitions $\mathcal{T}$. Places represent the conditions or states of the system, while transitions model the events or actions that cause state changes. The graph's edges, or arcs, represent the flow of tokens between places and transitions, specifying the conditions required for transitions to fire.

For any node $n \in \mathcal{P} \cup \mathcal{T}$, we define $\pi(n)$ as the set of input (upstream) nodes and $\sigma(n)$ as the set of output (downstream) nodes. The marking $\mu(p)$ represents the token distribution at place $p$. When a transition $t \in \mathcal{T}$ fires, the new marking is given by:

\begin{equation}
\Tilde \mu(p)=
    \begin{cases}
     \mu(p)-1 \quad \forall\ p \in \pi(t) \\
     \mu(p)+1 \quad \forall\ p \in \sigma(t) \\
     \mu(p)  \quad\text{otherwise}
    \end{cases}  
    \label{eq: arc expression }
\end{equation}
\vspace{5pt}

In addition to standard transitions, some transitions in a Petri Net may be timed transitions, which do not fire until a token has spent the necessary sojourn time in the upstream place. This property allows for modeling delays inherent to processes, such as machine operation times or transportation durations in FMS.

Colored Petri Nets enhance the traditional Petri Net model by allowing tokens to carry additional information, referred to as "colors", thus providing greater expressiveness. This feature enables the compact representation of processes with similar structures but differing properties, making CPNs particularly useful in job-shop scheduling problems, where multiple jobs utilize the same shop floor~\cite{Jensen.1992}. A marked colored Petri Net is formally defined by:

\begin{figure*}[ht]
\centering
    \includegraphics[width=1\linewidth]{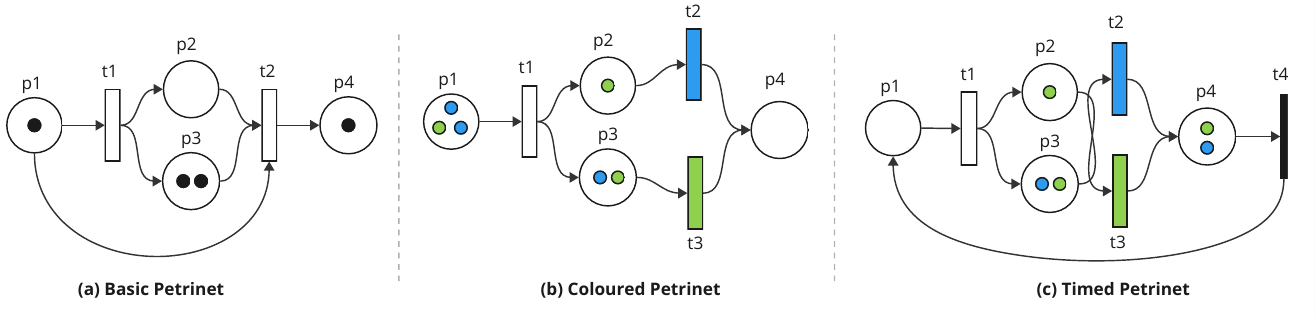}
\caption{Examples of Petri Net Variants: (a)  basic Petri Net, (b)  colored Petri Net, and (c)  colored-timed Petri Net.}
\label{fig: Petri Nets example}
\end{figure*}

\begin{equation}
\label{eq:cpn_definition}
\text{CPN} = (\mathcal{P}, \mathcal{T}, \mathcal{A}, \Sigma, \mathit{C}, \mathit{N}, \mathit{E}, \mathit{G}, \mathit{I})
\end{equation}
where:
\begin{itemize}[itemsep=0.2em]
     \item \(\mathcal{P} = \{ \mathit{p}_1, \ldots, \mathit{p}_m \}\): set of places,
    \item \(\mathcal{T} = \{ \mathit{t}_1, \ldots, \mathit{t}_n \}\): set of transitions,
    \item \(\mathcal{A} = \{ \mathit{a}_1, \ldots, \mathit{a}_k \}\): set of arcs,
    \item \(\Sigma = \{ \mathit{c}_1, \ldots, \mathit{c}_l \}\): set of colors ,
    \item \(\mathit{C} : \mathcal{P} \cup \mathcal{T} \rightarrow \Phi(\Sigma),\ \Phi(\Sigma) \subseteq \Sigma\): color set function,
    \item \(\mathit{N} : \mathcal{A} \rightarrow (\mathcal{P} \times \mathcal{T}) \cup (\mathcal{T} \times \mathcal{P})\): node connectivity function,
    \item \(\mathit{E} : \mathcal{A} \rightarrow \mathit{e}\): arc expression function,
    \item \(\mathit{G} : \mathcal{T} \rightarrow \{0, 1\}\): transition guard function,
    \item \(\mathit{I} : \mathcal{P} \rightarrow \text{initiation sequence}\): initialization function.
\end{itemize}
\vspace{5pt}

The Petri net is a powerful modeling tool used to represent various systems. In this paper, we use it to model a manufacturing system. At the heart of the model are \textbf{tokens}, which, among other things, represent operations to be performed on workpieces. A job in a JSSP is an ordered sequence of operations. Each token carries a \textbf{color} that encodes constraints, such as the type of operation, ensuring that it can only be processed on compatible machines. \textbf{Places} represent the states of system resources, for example, a place associated with a machine holds a token if the machine is currently busy, while other places may represent buffers or job queues. \textbf{Transitions} model the events or actions that change the system’s state. When a transition such as “allocate” fires, it moves a token from the buffer to the machine, indicating the start of a processing task. \textbf{Arcs} define the directional connections between places and transitions, representing the flow of tokens through the system. Each arc has an associated \textbf{expression} that specifies how many tokens are transferred when a transition fires. Since each token typically represents one operation, the arc expression defaults to one in our case. Transition \textbf{guard functions} are Boolean conditions that determine whether a transition is enabled. For example, an allocation transition is only enabled if there is a token in the buffer, the machine is idle, and the token’s color matches the machine’s capabilities. Lastly, the \textbf{initialization} function defines the system’s starting state, including which machines are occupied, the number of operations in queues and buffers, and the initial distribution of tokens. This model provides a structured and expressive way to simulate and analyze the dynamic behavior of manufacturing systems.

\bigskip

In Figure~\ref{fig: Petri Nets example}, we provide a simple example of the different  Petri Net variants: (a) Basic Petri Net illustrating transitions and places; (b) Colored Petri Net showing token differentiation with 'colors'; (c) Colored-Timed Petri Net, extending the model with time to control the firing delay for transitions.

In sub-figure (a), transition $t_1$ is enabled because its only upstream place contains a token. In contrast, transition $t_2$ is not enabled, as one of its upstream places, $p_2$, is empty. Therefore, $t_1$ must fire first for $t_2$ to become enabled.

In sub-figure (b), even though $p_2$, the upstream place of $t_2$, contains a token, transition $t_2$ is not enabled because the token's color does not match the transition's color. The color here is for simplicity; however, in actual implementation, it can be an attribute, a tuple, a vector, or another data structure that must meet the transition's requirements.

Finally, in sub-figure (c), a timed transition is added. This transition will not fire until the token has spent its sojourn time in the upstream place $p_4$. In our implementation, the sojourn time is included as part of the token's attributes to represent processing times, transport times, and similar delays.

\section{Problem and Modelling }
\label{section: Problem and Modelling}

This section extends the classical JSSP to incorporate material handling and tool sharing in an FMS. Unlike the traditional JSSP, this scenario introduces limited AGVs, constrained tool availability, and tool transporters, adding scheduling complexity. The objective is to minimize the makespan by efficiently coordinating machines, AGVs, tools, and transporters. We begin with the problem definition, key assumptions, and constraints.

\subsection{Problem Definitions}
 \label{section: problem definition}

\begin{figure}[ht]
\centering
\includegraphics[width=\linewidth]{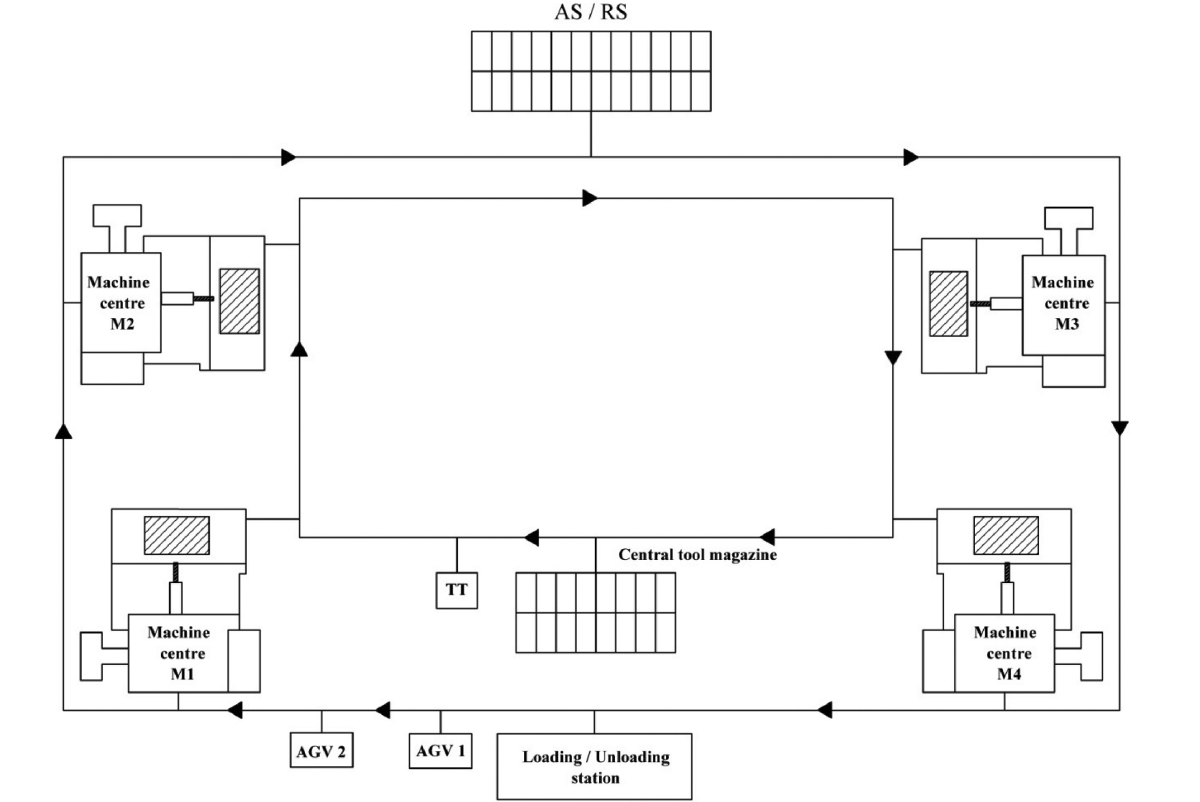}
\caption{An FSM problem example \cite{Reddy.2021}}
\label{fig: FMS}
\end{figure}

This problem extends the classical JSSP by incorporating material handling and tool management logistics within an FMS. In a traditional JSSP, each job consists of multiple operations that must be processed on specific machines. However, in this enhanced scenario, additional complexities arise due to the limited availability of AGVs for material transport, a restricted number of machines and tools, and a limited number of tool transporters that transport tools between machines. These constraints add logistical and scheduling layers that must be addressed for an efficient manufacturing process. The objective of the scheduling problem is to minimize the total time, or makespan, required to complete all jobs while considering machine, AGVs, tools, and tool transporters availability. Figure~\ref{fig: FMS} depicts an example of an FMS.

\subsection{Key Assumptions}

\begin{enumerate}
    \item \textbf{Job Structure:} Each job $J_i$ consists of a sequence of operations $O_{ij}$, where $O_{ij}$ denotes the $j^{\text{th}}$ operation in the $i^{\text{th}}$ job. Operations must follow a specific order, and each requires a machine $m \in M$, a tool $t \in T$, and a processing time $p \in \mathbb{Z}^+$.

    \item \textbf{Machine Processing:} Each operation $O_{ij}$ must be processed on a specific machine $m_k$. The processing time $p_{ij}$ is fixed. Each machine can process only one operation at a time.

    \item \textbf{Material Handling (AGVs):} Materials are transported by AGVs, defined as
    \begin{equation}
    \label{eq:AGVset}
    A = \{A_1, A_2, \ldots, A_q\}
    \end{equation}
    where $q$ is the number of AGVs. Each AGV handles one material transport at a time and cannot be reassigned until the current transport is completed.

    Travel times between machines are given by
    \begin{equation}
    \label{eq: DAGV}
    D_{AGV} = \bigl[d^{AGV}_{uv}\bigr]
    \end{equation}
    where $d^{AGV}_{uv}$ is the time to travel from machine $u$ to machine $v$. Deadheading (travel without load) is considered.

    \item \textbf{Tools and Tool Management:} Each operation $O_{ij}$ requires a specific tool $t_l \in T$. Only one instance per tool type exists. For the first operation, tools are fetched from the Tool Magazine using tool transporters:
    \begin{equation}
    \label{eq:TTset}
    TT = \{TT_1, TT_2, \ldots, TT_s\}
    \end{equation}
    where $s$ is the number of tool transporters. Transport times between machines are given by
    \begin{equation}
    \label{eq: DTT}
    D_{TT} = \bigl[d^{TT}_{uv}\bigr]
    \end{equation}
    where $d^{TT}_{uv}$ is the time to move a tool from machine $u$ to $v$. Deadheading is also considered.

    \item \textbf{System Layout:} The system includes machines, a Tool Magazine, and Load/Unload Stations. Locations influence transport times defined in \eqref{eq: DAGV} and \eqref{eq: DTT}.

    \item \textbf{Load/Unload and Tool Magazine:} Materials for the first operation are picked up from a Load/Unload Station by an AGV. Tools are retrieved from the Tool Magazine by a tool transporter.
\end{enumerate}

\subsection{Constraints}

\begin{enumerate}
    \item \textbf{Machine Availability:} Each machine $m_k$ can process only one operation at a time. Once an operation starts on a machine, no other operation can use that machine until the current operation finishes.

    \item \textbf{AGV Availability:} Only a limited number $q$ of AGVs are available for material transport. If all AGVs are in use, any job waiting for material transport will be delayed.

    \item \textbf{Tool Availability:} In addition to machine assignment, each operation may require a specific tool type $t_l$, which introduces tool availability as a further constraint. Tools are treated as limited shared resources: each type is assumed to be available as a single entity that can be used on different machines. If multiple operations require the same tool type, they must be scheduled to avoid conflicts. 
    \item \textbf{Tool Transporter Availability:} Similar to AGVs, the number of tool transporters is limited. If all $s$ transporters are in use, any job waiting for a tool will be delayed until a transporter becomes available.

    \item \textbf{Job precedence Constraints:} Operations follow an ordered sequence: $O_{ij}$ can begin only after $O_{ij-1}$ is completed and both the material and tool are available at the machine.

    \item \textbf{Deadheading:} If an AGV or tool transporter is not at the required location dictated by the scheduler to carry a piece for the next operation, it must travel empty to reposition itself. This unladen movement, referred to as \textit{deadheading}, is a non-productive time that is added to the total makespan.

\end{enumerate}

\begin{figure*}[ht]
\centering
\includegraphics[width=\linewidth]{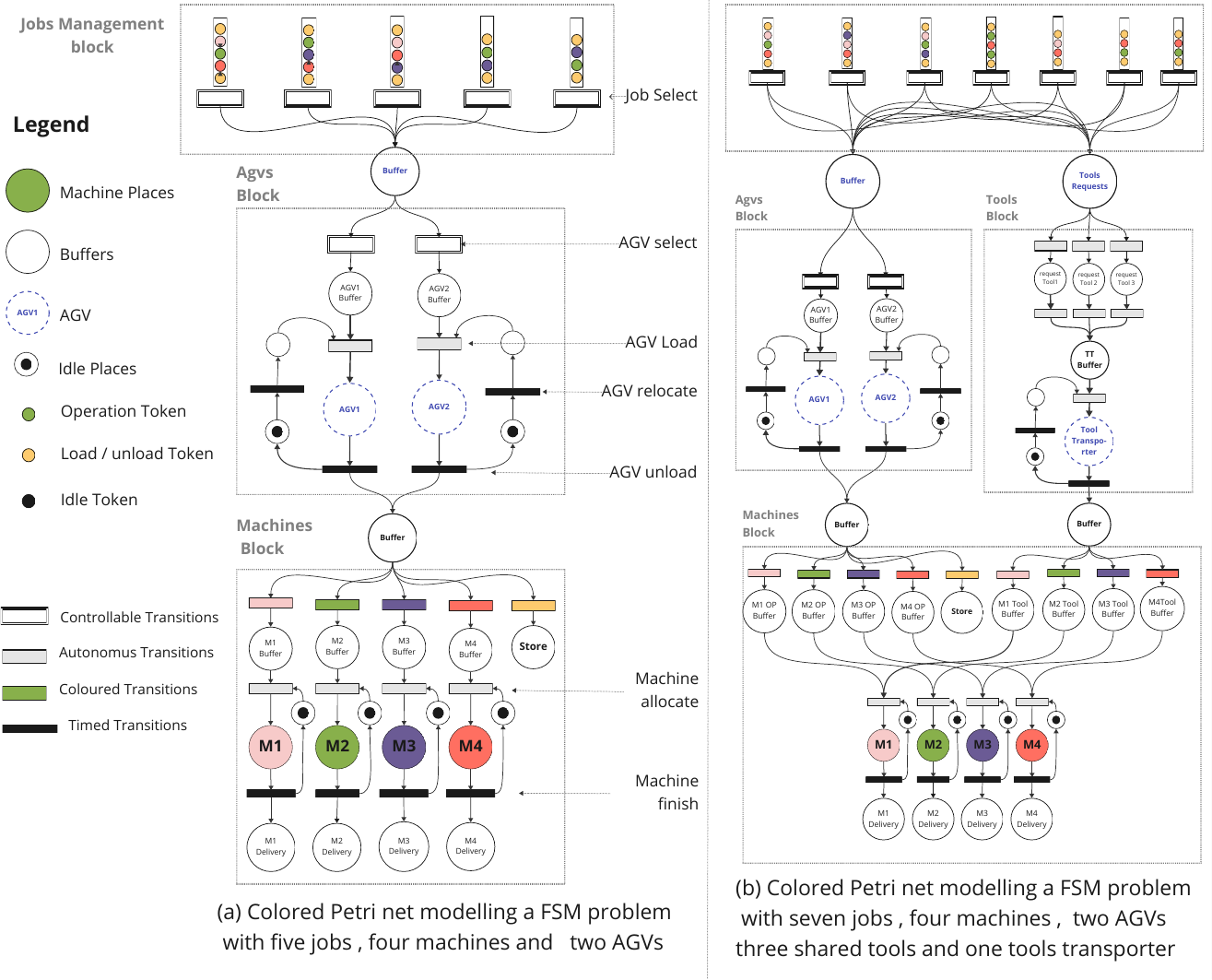}
\caption{\textbf{Colored-Timed Petri Nets modelling a FSM problems}}
\label{fig: petrinet}
\end{figure*}
\section{Integrating PetriRL for intralogistics }
\label{section: Methodology  and Implementation}

This section outlines our methodology and details the implementation. We will discuss the framework used, the design of the algorithms, and the specific implementations within our system.

\subsection{FMS Environment using Colored-Timed Petri Nets }

Building on our previous work in \cite{Lassoued.2024}, We model the FMS using a colored-timed Petri Net. Figure~\ref{fig: petrinet} illustrates the two study cases included in this research with increasing complexity. The first step depicted in Subfigure (a) focuses solely on the AGVs. Then, in Subfigure (b), we also integrate tool-sharing between the machine and the associated tool transporter.

Subfigure (a) illustrates the CTPN model of the FMS problem with five jobs, four machines, and two AGVs. The model comprises three main blocks: the jobs management block, the AGVs block, and the machines block. The jobs management block represents the jobs to be processed and their ordered list of operations, where each operation is depicted by a colored token, indicating the designated machine for processing. In addition to color, each token holds specific attributes related to the operation, such as processing time. These tokens are placed in job-related places that link to controllable transitions: "job selection." The sequence of transition firings determines the order in which the system will process the operations. An external trigger, in addition to the presence of a token, is needed to fire these controllable transitions. This trigger is managed by the RL agent, thus controlling the flow of tokens.

The AGVs block follows the jobs management block, where the RL agent makes a second decision: selecting the appropriate AGVs to transport the operation token. The remainder of this process is automated through autonomous and timed transitions, allowing the RL agent to focus on the key decisions of job prioritization and AGVs allocation. In this setup, the AGV load transition fires automatically when a token is present in the AGV buffer and the AGV is idle. Once an AGV is allocated, the transport time is managed by a timed transition (unload transition), which cannot fire until the token has remained for the required transport time from one machine to the next. Upon firing the unload transition, the token moves to the machine block. The AGV is repositioned for the next token in a process known as deadheading, in which the AGV travels empty. Deadheading introduces complexity, as the time needed for this empty travel cannot be predetermined without knowing the sequence of token processing and locating the next destination.

The final block, the machine's block, directs transported tokens to the appropriate machines using a color filter, where a colored transition ensures that each operation token reaches its designated machine. Allocation occurs automatically when a token is in the machine buffer and the machine is idle. The presence of one token in the machine's idle place guarantees that only one operation is processed at a time on each machine. Processing time is simulated with a timed transition. The machine's transition cannot fire until the token has spent the required sojourn time in the machine's processing place.

In sub-figure (b), a tool-sharing block is added. When a job selection transition fires, it generates two identical tokens: one follows the previously described path through the AGV, while the other represents a tool request for the operation. Each token is color-coded with a triplet (job, machine, tool), where the tool request token is sorted by the required tool type and directed to the tool transport system. The tool transport system has a structure similar to that of AGVs. A machine can only start processing in this setup when three conditions are met. The piece to be processed is waiting in the machine buffer; the needed tool is ready, and the machine is idle.

Our approach is modular and explainable, a significant advantage of our method. These characteristics stem from our use of Petri Nets. Modularity is achieved because each transition relies solely on its immediate upstream nodes, a property referred to as locality. Locality allows each component of the Petri Net to operate independently, supporting a modular approach. This design allowed us to integrate the tool transport block with minimal structural changes. Furthermore, this modularity facilitates smooth transitions between different configurations: we can easily switch from an AGV + Tool sharing model to an AGVs-only model or even to a standard JSSP model by simply removing the AGVs and tools blocks and connecting the job management block directly to the machines block.

While most existing approaches to solving  JSSP using RL rely on disjunctive graphs, our method emphasizes explainability by modeling FMS with a Petri net. Through token distribution, we graphically represent the system state, enabling real-time tracking of key aspects such as remaining operations, AGVs in transport or deadheading, idle machines, and completed tasks. The movement of tokens illustrates the interactions between machines and reveals system constraints, offering an intuitive understanding of dynamics and dependencies. This enhances transparency and supports informed, real-time decision-making.

\subsection{RL environment and Maskable PPO}
\label{section: environment}

In this section, we define key components of our framework PetriRL for intralogistics, including the environment, agent, observations, action space, and reward function. The central elements of the environment are the Petri Net model of the FMS alongside the reward function.

In the Petri Net, some transitions are controllable. Namely, the job selects and AGV selects transitions, allowing the RL agent to effectively control them by providing the necessary external triggers. This enables the agent to manage the token flow without overseeing every transition, defining its action space. For instance, with 15 jobs and two AGVs, the agent has 17 possible actions, focusing on critical decisions like job operations and AGV allocation. Meanwhile, the Petri Net autonomously handles constraint enforcement and token management. This collaboration streamlines the system, reducing the RL agent's search space and enhancing overall efficiency.

Observations are derived from the token distribution within the Petri Net. At its simplest, this can be represented by the Petri Net marking, which indicates the number of tokens in each place. In our implementation, we enhance observations by incorporating token features such as remaining processing time and color, providing a richer, more informative view of the system's state for the RL agent.

Our primary goal is to minimize the makespan. However, using it as a reward poses challenges due to its sparsity, as feedback is only provided at the end of the episode. While employing makespan as a reward has been effective in small instances where the number of actions before termination is limited, it becomes problematic in larger instances. In these cases, the lengthy action sequences create a credit assignment problem, making it challenging to identify which actions positively or negatively influenced the final makespan.

We penalize idle machines to address sparse rewards in larger instances, providing immediate feedback through machine state changes. This strategy penalizes delays in material and tool availability, optimizing AGVs and tool transporters' schedules. By focusing on penalizing idleness instead of rewarding active use, the agent learns to complete jobs more efficiently, minimizing time and incurring penalties. The reward function is given by

\begin{equation}
\label{equation: reward function}
\text{Reward} =  - \left( \frac{\sum \text{idle machines}}{\text{total number of machines}} \right).
\end{equation}

\bigskip

We selected Maskable Proximal Policy Optimization from the Stable Baselines 3 library due to its stability, generalization capabilities, and suitability for environments with complex action spaces and dynamic constraints. The maskable variant allows the agent to dynamically restrict its action space based on the current state, improving performance and efficiency. This dynamic masking is driven by the Petri Net’s guard function, which generates a Boolean vector indicating enabled transitions based on token distribution. The agent uses this mask to select only feasible actions, significantly reducing the search space and computational overhead. The novelty of our approach lies in the integration of traditional Petri nets with modern deep reinforcement learning. Extending our previous work on classical Job Shop Scheduling Problems \cite{Lassoued.2024} , we now address additional intralogistics challenges such as AGV coordination and tool sharing. The Petri net provides a scalable and modular modeling framework, and its guard functions synergize with PPO's action masking to further enhance computational efficiency. The corresponding pseudocode is provided in Algorithm \ref{algorithm: maskable PPO}.

\bigskip

\noindent\textbf{Variables Used in Algorithm~\ref{algorithm: maskable PPO}:}
\vspace{-1mm}
\begin{itemize}[label=-, itemsep=1pt, parsep=0pt]
  \item $\hat{A}_t$: Estimated advantage function at time step $t$.
  \item $R_t$: Discounted cumulative return at time step $t$.
  \item $V_{\theta}(s_t)$: Value of state $s_t$ under the current policy.
  \item $V_{\theta_{\text{old}}}(s_t)$: Value of state $s_t$ from the previous iteration.
  \item $G(a_t)$: Petri Net guard function.
  \item $z_t(a_t)$: Logit  for action $a_t$ at time $t$.
  \item $\pi_\theta(a_t | s_t)$: Probability of selecting $a_t$ in state $s_t$ .
  \item $\pi_{\theta_{\text{old}}}(a_t | s_t)$: Probability of selecting $a_t$ in state $s_t$ under previous policy.
  \item $\rho_t$: Importance sampling ratio: $\rho_t = \frac{\pi_\theta(a_t | s_t)}{\pi_{\theta_{\text{old}}}(a_t | s_t)}$.
  \item $\epsilon$: Clipping parameter for PPO's policy update.
  \item $L^{VF}(\theta)$: Value function loss term.
  \item $L^{CLIP}(\theta)$: Clipped objective .
  \item $L(\theta)$: Final PPO loss combining policy and value terms.
  \item $C_1$: Coefficient balancing the value function loss.
  \item $\alpha$: Learning rate for policy updates.
\end{itemize}
\vspace{1mm}

\begin{algorithm}[ht]
  \small
  \caption{Proximal Policy Optimization with Action Masking using Petri Nets' guard function \cite{Lassoued.2024}.}
  \label{algorithm: maskable PPO}
    \begin{algorithmic}[1]
      \State Initialize policy parameters $\theta$, value function parameters $\phi$
      \State Set hyperparameters, including the clipping parameter $\epsilon$
      \For {iteration $= 1, 2, \ldots$}
        \State Collect trajectories using the current policy $\pi_\theta$
        \State Compute advantages $\hat{A}_t$ and returns $R_t$
        \Statex \textbf{Update the Value Function:}
        \ state Compute the value function loss:
         \begin{align*}
          L^{VF}(\theta) = \mathbb{E} \left[ \left( V_{\theta_{\text{old}}} (s_t) - V_{\theta} (s_t) \right)^2 \right]
        \end{align*}
        \ state Optimize the value function parameters $\theta$
        \ state Compute the Mask using the Petri Net's guard function:
        \[ \text{Mask}[i] = G(a_i) \quad \forall a_i \in \mathcal{T}_a\]
        \ state Compute the actor's policy logits:
        \begin{equation*}
          z_t(a_t) =
          \begin{cases}
            \begin{aligned}
              &\text{policy\_network}(s_t, a_t) & \text{if} \ G(a_t)=1 \\
              &-\infty & \text{else} 
            \end{aligned}
          \end{cases}
        \end{equation*}
        \ state Convert the logits into a probability distribution over actions and calculate the action probability ratio:
        \begin{align*}
         \pi_{\theta}(a_t | s_t) &= \sum_{a'} \frac{e^{z_t(a)}}{e^{z_t(a')}} \qquad, \rho_t = \frac{\pi_\theta(a_t | s_t)}{\pi_{\theta_{\text{old}}}(a_t | s_t)}
        \end{align*}
        \State Compute Clipped surrogate objective:
        \begin{align*}
          L^{CLIP}(\theta) & = \mathbb{E}_t\left[\min\left(\rho_t \hat{A}_t, \text{clip}(\rho_t, 1 - \epsilon, 1 + \epsilon) \hat{A}_t\right)\right]
        \end{align*}
        \State The final PPO loss ($C_1$ is a balancing hyper-parameter) :
        \begin{align*}
          L(\theta) = L^{CLIP}(\theta) + C_1 L^{VF}(\theta) 
        \end{align*}
        \State Optimize the policy parameters $\theta$ ($\alpha$ is the learning rate):
        \begin{align*}
          \theta & \leftarrow \theta - \alpha \nabla_\theta L(\theta)
        \end{align*}
      \EndFor
    \end{algorithmic}
\end{algorithm}

\vspace{0.5em}

\subsection{AGVs transport time calculation and lookaheads}
\label{section: lookahead}

FMS introduces unique challenges compared to the traditional JSSP. In JSSP, all processing times are known in advance. In FMS, however, the transport times of AGVs depend on the processing sequence and thus can't be predetermined until this sequence is known. While the loaded travel time within a job is fixed, the fact that AGVs are shared across jobs introduces additional times for empty travel, called "deadheading," which must be added to the total makespan.

In our environment, AGVs transport times and deadheading are calculated using the transport time matrix \( D_{\text{AGV}} \) introduced in Section \ref{section: problem definition}. Once an "AGV's unload" transition fires, it relocates to the next position, determined by the next token in its buffer. The relocation time is managed by a timed relocation transition, ensuring the AGV spends the required time moving from its current position to the next. If no relocation is needed, the time is zero, as the diagonals of \( D_{\text{AGV}} \) are zero.

A problem arises when the AGV buffer is empty at the time step when the relocation time needs to be calculated. To address this challenge, we opted to assign the maximum transport time in \( D_{\text{AGV}} \) if the next position is unknown. This approach has two benefits. First, the RL agent will be incentivized to fill the buffer whenever possible to minimize the total makespan. Second, it ensures a worst-case but correct plan, as the maximum time is allocated if the destination is yet to be determined.

This problem can be further optimized by predicting the token that will fill the buffer in the future. Inspired by the simulation phase of the Dyna-Q algorithm \cite{Sutton.2018}, we leverage the benefits of a model-based RL approach. In this approach, a lookahead mechanism anticipates future states of the AGV buffers and optimizes its transport times accordingly. We highlight two possible scenarios:

\begin{enumerate}
    \item \textbf{Scenario 1} If a token is present in the buffer, the transport time is calculated using the transport time matrix \( D_{\text{AGV}} \) based on the current position and the next token's operation collection position.

     \item \textbf{Scenario 2} If the buffer is empty, the algorithm creates a twin of the environment and simulates future steps using the learned current policy until the buffer is filled with a token. The transport time is then calculated as in scenario one.
     
\end{enumerate}

Given the deterministic nature of the current RL policy, if an AGV buffer is empty and operations remain in the job queue, we can simulate future steps to accurately predict which token will be assigned to each AGV. This foresight enables the AGV to proactively position itself at the appropriate machine, ready to transport the corresponding piece to its next processing step. By ensuring that the AGV is optimally positioned for the upcoming token, deadheading is effectively eliminated, thereby contributing to a reduced makespan. The ablation study further analyzes the benefits of this lookahead strategy, underscoring its role in enhancing overall system performance.
 
\section{Results and Discussion}
\label{section: Results and Discussion}

In this section, we present the results of our experiments, beginning with introducing a new benchmark for large instances to evaluate the performance of our approach. We compare our method against heuristics and metaheuristics, detailing the training processes involved. We then analyze the experimental results, examining our solution's scalability and dynamic behavior. Finally, we conduct an ablation study to assess the contributions of various components within our model.

\subsection{Large instance benchmark}
\label{section: new benchmark}

Our literature review identified a gap in existing research: most studies integrating AGVs and tool transport rely on the Bilge benchmark\cite{Bilge.1995}, which provides problem sets with various layouts to evaluate AGVs performance. However, a key limitation of this benchmark is that the scenarios typically involve no more than eight jobs and six machines. Real-world applications, however, often demand larger configurations with more jobs and machines. We propose a new benchmark based on the widely recognized Taillard benchmark to address this gap. Our benchmark is compatible with standard JSSP problems, featuring instances with up to 100 jobs and 20 machines while incorporating tool sharing and AGVs transport times. The increased complexity of this benchmark makes it impractical for exact methods due to the expanded search space, necessitating the use of better adapted optimization techniques such as heuristics, metaheuristics, and reinforcement learning.

The new benchmark consists of 80 instances, organized into eight groups with increasing complexity, ranging from 15x15x15 to 100x20x20  (jobs x machines x tools). Each group contains ten instances of the same configuration, populated with random values generated using the Linear Congruential Generator (LCG), a pseudo-random generator. The pseudocode for LCG is presented in Algorithm \ref{alg:instance_generator}. To ensure the reproducibility of our results, we also provide the Table \ref{table: seeds} containing the seeds used to generate these instances.

\begin{algorithm}[ht]
\caption{Random instances generator using LCG \cite{Taillard.1993}}
\label{alg:instance_generator}
\begin{algorithmic}[1]
    \State \textbf{Linear Congruential Generator (LCG):}
    \State Constants: $a = 16807$, $b = 127773$, $c = 2836$, $m = 2^{31} - 1$
    \State Calculate $k = \left\lfloor \frac{{X_i}}{b} \right\rfloor$
    \State Update the seed: $\text{X}_{i+1} = a \cdot (\text{X}_i \mod b) - k \cdot c$
    \If{$\text{X}_{i+1} < 0$}
        \State Let $\text{X}_{i+1} = \text{X}_{i+1} + m$
    \EndIf
    \State Generate the pseudorandom number: $u(0,1) = \frac{\text{X}_{i+1}}{m}$
    \State Generate the random integer in the range $[a, b]$: $U[a, b] = [a + u(0,1) \cdot (b - a + 1)]$
    \State \textbf{Return} $U[a, b]$
    
    \State \textbf{Generate the processing time using (use time-seed):}
    \State Initialize $d_{ij}$ for all $i$ and $j$: $d_{ij} \gets U[1, 99]$
    \For{$i = 1$ to $n$}
        \For{$j = 1$ to $m$}
            \State $M_{ij} \gets j$
        \EndFor
    \EndFor
    \State \textbf{Generate the machining sequence (use machine-seed):}
    \For{$i = 1$ to $n$}
        \For{$j = 1$ to $m$}
            \State Swap $M_{ij}$ and $M_{iU[j,m]}$
        \EndFor
    \EndFor
\end{algorithmic}
\end{algorithm}

\begin{table}[H]
  \centering
      \resizebox{\linewidth}{!}{%
    \begin{tabular}{p{6cm}p{4cm}}  
      \toprule
      \textbf{Parameter} & \textbf{Seed Value} \\
      \midrule
      Machine Allocation Seed        & 398197754 \\
      Tool Allocation Seed           & 170719940 \\
      Processing Times Seed          & 840612802 \\
      Tools Transport Times Seed     & 280219920 \\
      AGVs Transport Times Seed      & 180119550 \\
      \bottomrule
    \end{tabular}%
    }
  \caption{Seed values used for generating the large-instance benchmarks.}
  \label{table: seeds}
\end{table}

\subsection{Competing algorithms}

After defining the problem environment and selecting the benchmark, we introduce the competing algorithms and the evaluation benchmarks. The following sections provide an overview of heuristic methods and a replication of the  Symbiotic Organisms Search (SOS) algorithm.

\subsubsection{Heuristics}
\label{section:heuristics}

Heuristics are practical methods for finding near-optimal solutions quickly. They simplify complex scheduling and resource allocation problems by applying predefined rules. While they don't guarantee the best solution, they provide good enough results in less time. However, they often require extensive domain knowledge.

In the context of FMS, we define two families of heuristics: the first is applied to selecting which jobs to process, and the second focuses on choosing the Automated Guided Vehicle to transport the jobs. Table \ref{tab:heuristics} lists the job selection heuristics that were implemented.

\begin{table}[ht]
  \centering
    \resizebox{\linewidth}{!}{%
    \begin{tabular}{p{1.3cm}p{9cm}}
      \toprule
      \multicolumn{2}{c}{\textbf{Job Selection Heuristics}} \\
      \midrule
      FIFO  & Job that entered the system earliest \\
      SPS   & Job with the shortest processing sequence  \\
      LPS   & Job with the longest processing sequence \\
      SPTN  & Job with the shortest time of the next operation \\
      LPTN  & Job with the longest time of the next operation \\
      MTWR  & Job with the most total work remaining \\
      LTWR  & Job with the least total work remaining \\
      LWT   & Job with the longest waiting time \\
      SPT   & Job with the shortest total processing time \\
      LPT   & Job with the longest total processing time \\
      SPSR  & Job with the shortest remaining processing sequence \\
      LPSR  & Job with the longest remaining processing sequence \\

      \bottomrule
    \end{tabular}%

    }
  
  \caption{List of heuristics used for benchmarking.}
  \label{tab:heuristics}
\end{table}

\subsubsection{Metaheuristics: symbiotic organisms search (SOS) }
\label{section: SOS replication}

\begin{figure}[ht]
\centering
\includegraphics[width=\linewidth]{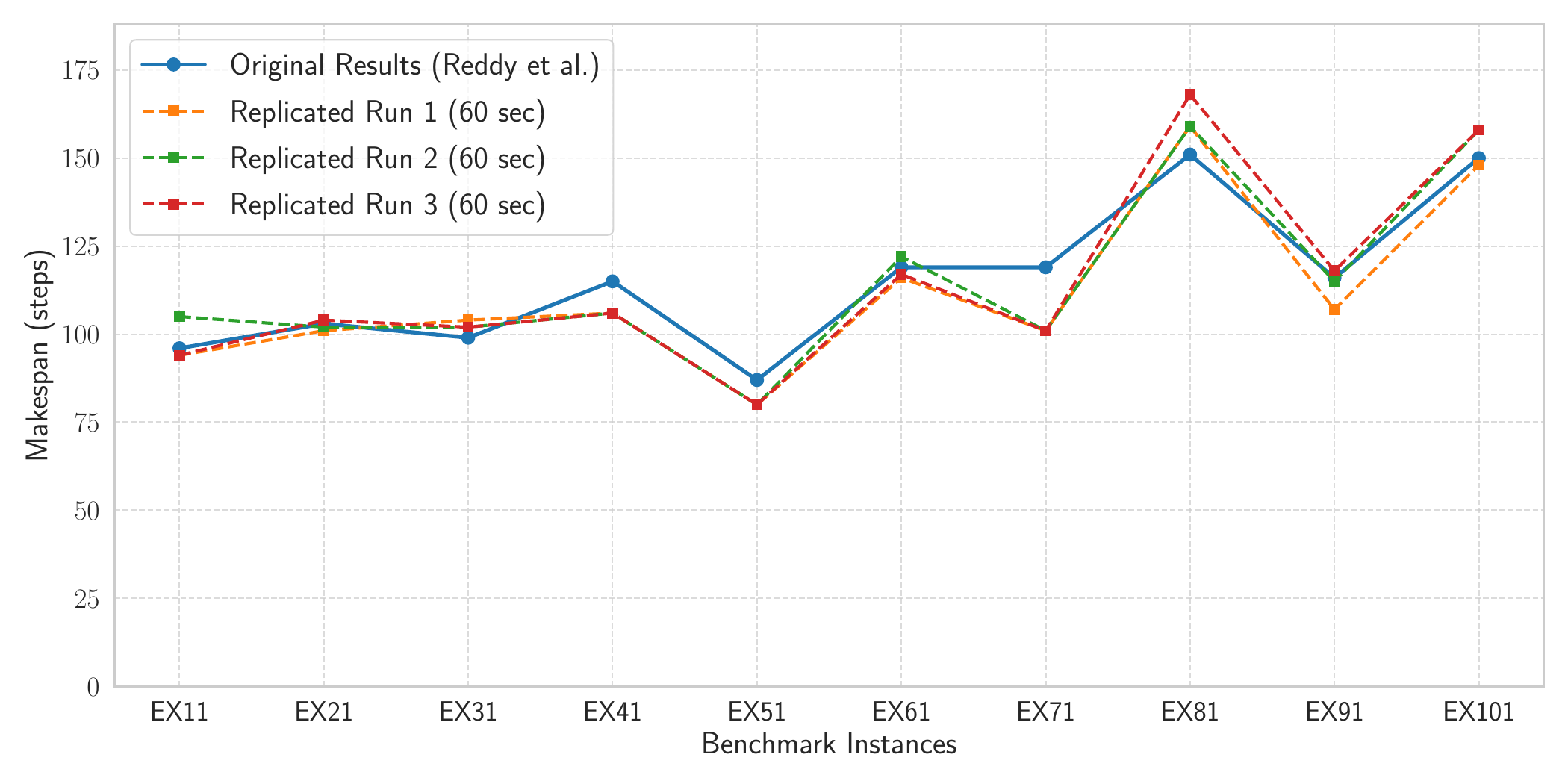}
\caption{Evaluation of Replicated Algorithm Performance Against the Original Implementation. Results from three runs of the replicated algorithm are compared to the original results to account for randomness.}

\label{fig: replication}
\end{figure}

Many metaheuristics have been explored to solve scheduling problems, but few have integrated AGVs. Our literature review identified only one study that addressed JSSPs with simultaneously AGVs and tool sharing, utilizing Symbiotic Organism Search (SOS) \cite{Reddy.2021}. Metaheuristics are known for enhancing search capabilities, and SOS takes this further by mimicking natural symbiosis. Its ability to facilitate cooperation among solutions and share beneficial traits helps avoid local optima and improve solution quality. While the original study demonstrated the effectiveness of SOS, its implementation was restricted to small instances. We replicated the provided algorithm in the original paper to evaluate its potential for larger-scale problems. Figure~\ref{fig: replication} illustrates the performance of our replication compared to the original results, confirming the accuracy and fidelity of our implementation.

As a search-based optimization algorithm, Symbiotic Organisms Search excels at exploring the entire search space and uncovering diverse solutions. However, increased complexity may limit its performance in larger search spaces. Additionally, SOS lacks knowledge retention, necessitating a complete re-execution of the search with each change, which diminishes its adaptability in dynamic environments. These observations will be validated in the experimental results and analysis section.

To ensure a fair comparison with our RL agent, the heuristics and metaheuristics interact with the same Petri net environment. All methods operate under identical constraints and optimization objectives. At each time step, the environment produces a Boolean list of enabled actions, determined by the guard functions. This list is shared with all the RL agent, the heuristic, and the metaheuristic methods, allowing each algorithm to select the most appropriate action from the same action space. The scheduling process continues iteratively until all jobs are completed, after which the resulting makespan is compared across methods to evaluate their performance.

\subsection{Training}
The experiments were conducted on a machine equipped with an NVIDIA Quadro RTX 5000. The models were implemented using the PyTorch deep learning framework and the Stable Baselines 3 library \cite{stable-baselines3}, and all experiments were performed on a system running Windows 11. The number of training steps, the training time, and the deployment time are reported in Figure~\ref{fig: time_log}:

\begin{figure}[ht]
\centering
\includegraphics[width=\linewidth]{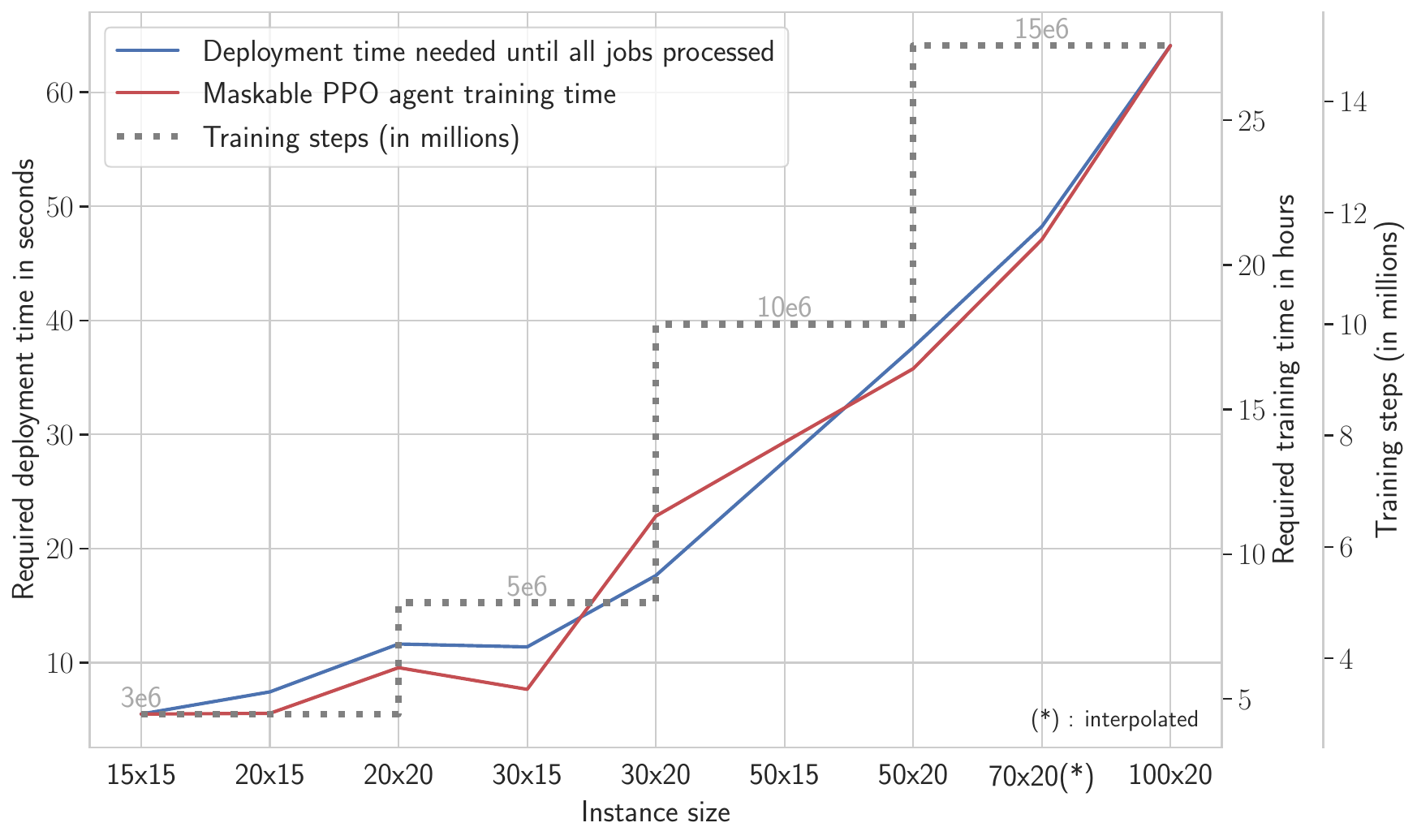}
\caption{Training and deployment time for different instances.
The red line represents the agent training time requirement for different instance sizes, as shown on the corresponding y-axis on the right.
The blue line represents the time required to solve the FSM problem during the inference. The corresponding y-axis is on the left.
The dotted grey line represents the increasing number of training steps in every size group.}
\label{fig: time_log}
\end{figure}

\begin{figure}[ht]
\centering
\includegraphics[width=\linewidth]{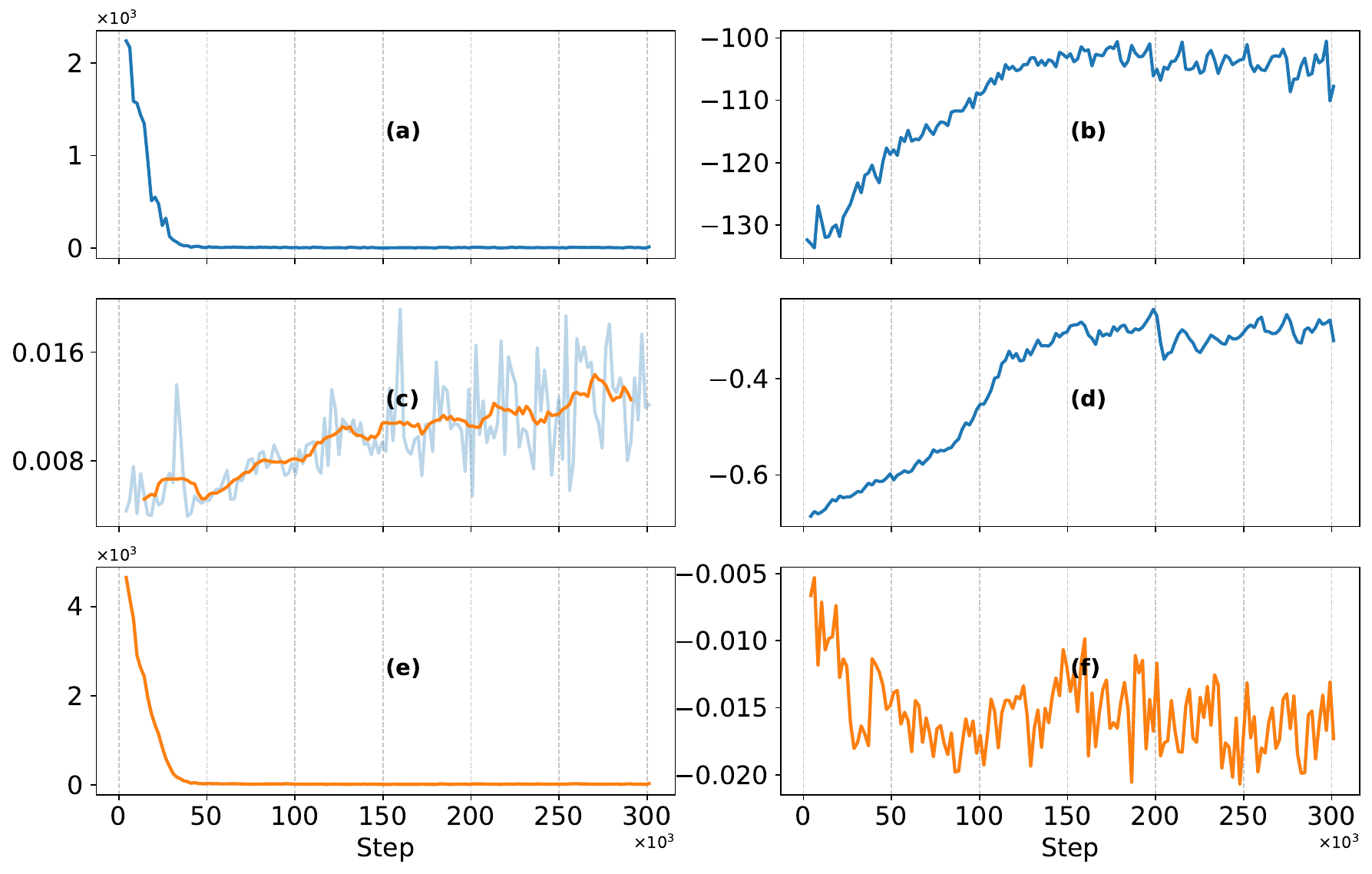}
\caption{Agent training performances, for the job set 1 layout 1, 5 jobs x4 machines x2 AGVs. (\textbf{a}) the combined training loss, (\textbf{b}) the episode mean reward,(\textbf{c}) the Kullback–Leibler divergence evolution, (\textbf{d}) the entropy loss,(\textbf{e}) the value function loss, (\textbf{f}) the policy loss. The agent is trained for \textbf{3e5} steps using the maskable PPO algorithm.}
\label{fig: training}
\end{figure}

Six metrics were selected to analyze the training process depicted in Figure~\ref{fig: training}: the combined training loss, the episode mean reward, approximate \textit{KL} divergence, entropy loss, value function loss, and policy loss. The episode mean length and reward indicate overall training performance, while \textit{KL} divergence is used to assess training stability, and entropy loss evaluates the exploration-exploitation balance.

\begin{table*}[ht]
  \centering
  \resizebox{\linewidth}{!}{%
  \begin{tabular}{p{1.2cm}p{1cm}p{1cm}p{1cm}p{1cm}p{1.5cm}|p{1cm}p{1cm}p{1cm}p{1cm}p{1cm}p{1cm}p{1cm}p{1cm}p{1cm}p{1cm}p{1cm}p{1cm}} 
    \toprule

    & \multicolumn{5}{c|}{} 
    & \multicolumn{12}{c}{\textbf{ transporter selection rule: Chooses the first AGV available}} \\

    \cmidrule(lr){2-6} \cmidrule(lr){7-18}

    Instance & Size & BU & Win & Reddy & PetriRL & FIFO & SPSR & LPSR & LTWR & MTWR & STPN & LPTN & LWT & SPT & LPT & SPS & LPS \\ 

    \midrule

    EX12   & 5x4  & 85  & 82  & 82  & 82  & 102 & 104 & 102 & 110 & 96  & 108 & 108 & 102 & 110 & 96  & 104 & 102 \\
    EX22   & 6x4  & 80  & 80  & 76  & 84  & 109 & 109 & 106 & 109 & 116 & 109 & 109 & 109 & 109 & 116 & 109 & 106 \\
    EX32   & 6x4  & 90  & 88  & 85  & 91  & 119 & 119 & 129 & 139 & 123 & 98  & 124 & 119 & 139 & 123 & 119 & 129 \\
    EX42   & 5x4  & 103 & 93  & 87  & 95  & 128 & 128 & 124 & 138 & 128 & 128 & 112 & 128 & 138 & 128 & 128 & 124 \\
    EX52   & 5x4  & 88  & 69  & 69  & 72  & 99  & 97  & 99  & 95  & 87  & 97  & 99  & 99  & 95  & 87  & 97  & 99  \\
    EX62   & 6x4  & 102 & 100 & 98  & 104 & 116 & 116 & 116 & 124 & 126 & 124 & 126 & 116 & 124 & 126 & 116 & 116 \\
    EX72   & 8x4  & 90  & 90  & 86  & 92  & 120 & 120 & 121 & 119 & 126 & 126 & 125 & 120 & 119 & 126 & 120 & 121 \\
    EX82   & 6x4  & 151 & 151 & 141 & 151 & 170 & 170 & 170 & 170 & 170 & 170 & 170 & 170 & 170 & 170 & 170 & 170 \\
    EX92   & 5x4  & 112 & 104 & 102 & 105 & 126 & 127 & 134 & 130 & 138 & 132 & 134 & 126 & 129 & 138 & 127 & 134 \\
    EX102  & 6x4  & 143 & 144 & 140 & 143 & 162 & 171 & 169 & 177 & 165 & 183 & 169 & 162 & 177 & 165 & 171 & 169 \\

    \midrule

    \textbf{Average} & & \textbf{104.4} & \textbf{100.1} & \textbf{96.6} & \textbf{101.9} 
    & 125.1 & 126.1 & 127.0 & 131.1 & 127.5 & 127.5 & 127.6 & 125.1 & 131.0 & 127.5 & 126.1 & 127.0 \\

    & \multicolumn{5}{c}{} 
    & \multicolumn{12}{c}{\textbf{ transporter selection rule: Chooses the AGV with least work done}} \\
       \cmidrule(lr){2-6} \cmidrule(lr){7-18}

    EX12   & 5x4  & 85  & 82  & 82  & 82  & 98  & 102 & 98  & 106 & 96  & 112 & 98  & 98  & 106 & 96  & 102 & 98 \\
    EX22   & 6x4  & 80  & 80  & 76  & 84  & 95  & 95  & 116 & 95  & 116 & 95  & 95  & 95  & 95  & 116 & 95  & 116 \\
    EX32   & 6x4  & 90  & 88  & 85  & 91  & 113 & 113 & 127 & 120 & 111 & 106 & 125 & 113 & 120 & 111 & 113 & 127 \\
    EX42   & 5x4  & 103 & 93  & 87  & 95  & 138 & 138 & 130 & 134 & 126 & 120 & 138 & 138 & 134 & 126 & 138 & 130 \\
    EX52   & 5x4  & 88  & 69  & 69  & 72  & 115 & 99  & 115 & 103 & 98  & 99  & 94  & 115 & 103 & 98  & 99  & 115 \\
    EX62   & 6x4  & 102 & 100 & 98  & 104 & 130 & 130 & 130 & 120 & 121 & 120 & 121 & 130 & 120 & 121 & 130 & 130 \\
    EX72   & 8x4  & 90  & 90  & 86  & 92  & 120 & 120 & 140 & 122 & 143 & 134 & 112 & 120 & 122 & 143 & 120 & 140 \\
    EX82   & 6x4  & 151 & 151 & 141 & 151 & 170 & 170 & 180 & 170 & 180 & 180 & 170 & 170 & 170 & 180 & 170 & 180 \\
    EX92   & 5x4  & 112 & 104 & 102 & 105 & 128 & 127 & 133 & 123 & 130 & 132 & 132 & 128 & 126 & 130 & 127 & 133 \\
    EX102  & 6x4  & 143 & 144 & 140 & 143 & 159 & 175 & 169 & 173 & 173 & 174 & 165 & 159 & 173 & 175 & 175 & 169 \\

    \midrule

    \textbf{Average} 
           &      & \textbf{104.4} & \textbf{100.1} & \textbf{96.6} & \textbf{101.9} 
           & 126.6 & 126.9 & 133.8 & 126.6 & 129.4 & 127.2 & 125.0 & 126.6 & 126.9 & 129.6 & 126.9 & 133.8 \\

    \bottomrule
  \end{tabular}
  }
  \caption{Comparison of Makespan using PetriRL versus various heuristics for job selection (FIFO, SPSR, LPSR, LTWR, MTWR, STPN, and LPTN, LWT, SPT, LPT, SPS, LPS) and Two AGV selection rules:   Chooses the first AGV available/  Chooses the AGV with least work done.}
  \label{table:Results_heuristics}
\end{table*}

In sub-figure (a), we observe a steady, monotonic decline in combined training loss without plateaus, signaling that the agent is effectively learning to solve the FSM problem. Concurrently, sub-figure (b) shows a consistent rise in mean episode reward, confirming that the training process enhances reward collection or, in this case, reduces penalties by promoting more efficient machine utilization through an optimized policy. These trends indicate that the agent is progressively refining its approach to improve performance and resource allocation. The slight volatility observed toward the end of subplot (b) is due to the exploration-exploitation trade-off: even after nearing convergence, the agent continues limited exploration to avoid local optima and improve policy robustness.

The approximate \textit{KL} divergence measures the difference between the new and old policies, assessing a key property of the proximal policy optimization algorithm, ensuring that policy updates are not overly drastic to maintain stability. In sub-figure (c), the \textit{KL}-loss is gradually rising, a sign of stable learning, ensuring that the policy evolves steadily and avoids drastic shifts.

Entropy loss quantifies the policy's randomness. High entropy indicates exploration, while low entropy suggests exploitation. The steady rise in entropy loss observed in sub-figure (d) throughout the training process indicates that the policy is becoming increasingly deterministic as the agent shifts toward exploitation.

The value function loss and overall loss metrics reflect training performance. In an actor-critic setup, the value function assesses the critic's ability to predict state values, which are essential for calculating action advantages used to update the actor network's parameters. Sub-figures (e) and (f) illustrate a monotonically decreasing value function loss, confirming that the critic effectively predicts state values, positively impacting the actor's ability to take reward-maximizing actions.

In summary, the training graphs reveal steady improvements: combined training loss decreases, and episode rewards rise monotonically. The increasing entropy loss reflects a shift to a more deterministic policy, while the actor and critic losses consistently decline, confirming effective learning.

\subsection{Experimental results and analysis}
After laying down the theoretical foundation, establishing the methodology to follow, and designating our benchmark, we proceed in this section with the experimental results and analysis. We then explore our approach's dynamic behavior and generalization capabilities, two key advantages of our approach. Finally, we will conduct an ablation study to assess the contributions of our model's different components.

\subsubsection{AGVs only implementation on small instances }

We start by testing our approach on implementations involving AGVs only, focusing on small problem instances. This choice is motivated by three main factors. First, it allows for a gradual increase in complexity, providing a structured approach to scaling the problem. Second, the initial research introducing AGVs into scheduling problems \cite{Bilge.1995} conducted its experiments exclusively with AGVs setups and small instance sizes. We can directly compare our results to this established benchmark by starting with similar conditions. Finally, testing on smaller instances serves as an initial filter, helping us identify the most effective approaches, which will be further evaluated along with our approach in the next chapter on larger instances and tool sharing.

Our approach was evaluated against heuristic and metaheuristic methods, with a detailed comparison to heuristics provided in Table \ref{table:Results_heuristics}. We compare our results against the twelve heuristics, "FIFO","SPSR," "LPSR," "LTWR," "MTWR," "STPN", "LPTN", "LWT", "SPT", "LPT", "SPS ", and "LPS" from  \cite{Kaban.2012}. Unlike traditional JSSPs, FMS require two types of heuristics: one for selecting jobs and another for allocating AGVs. In our analysis, we tested two distinct AGVs allocation strategies. The first strategy selects the first available AGV, maximizing its productivity and minimizing the number of AGVs required to serve all the jobs. The second strategy balances the workload across all AGVs, promoting even utilization and potentially extending AGVs' operational lifespans.

\begin{figure}[ht]
\centering
\includegraphics[width=\linewidth]{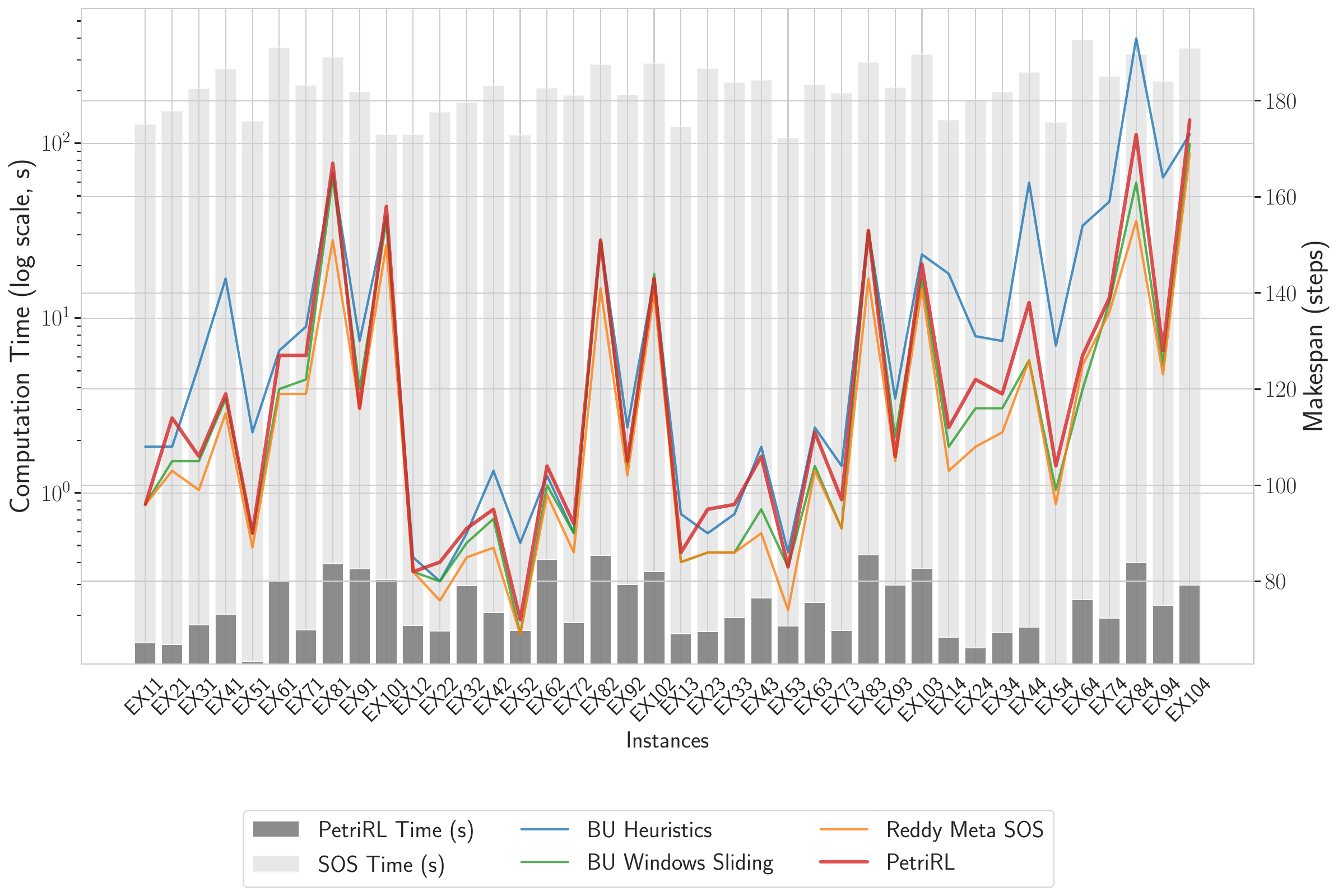}
\caption{Performance comparison across various instances using heuristics, metaheuristics, and our approach, PetriRL. The line plots show makespan results for each method, while the bar plot compares the computation time between our method and SOS on a logarithmic scale.}
\label{fig: benchmark_small}
\end{figure}

The results in Table \ref{table:Results_heuristics} demonstrate that our approach outperformed the tested heuristics with both proposed AGVs allocation strategies. This performance edge is mainly due to the flexibility of our method compared to rule-based heuristics. While heuristics are efficient in their simplicity, their rigid, predefined rules can limit performance, especially in AGV allocation. For example, the RL policies have the advantage of selecting AGVs based on factors such as their proximity to a target machine, which helps to minimize deadheading. In contrast, heuristics are rigid. For example, in the load-balanced strategy, the heuristics allocate the AGV with the least cumulative work regardless of its location, potentially leading to inefficiencies. We also note that the SOS metaheuristic is the best-performing algorithm on small instances. For a more detailed analysis, we present a comparative plot of our results against the SOS algorithm, among others, in Figure~\ref{fig: benchmark_small}.

In Fig~\ref{fig: benchmark_small}, we compare the results of our approach against the algorithms in \cite{Reddy.2021} and \cite{Bilge.1995}, including the sliding window approach and the SOS metaheuristic. The plot features two axes: one for makespan and another on a log scale to compare the computation time of our approach with the SOS metaheuristic. First, our approach demonstrates competitive performance, with results similar to the best alternatives and a slight advantage over SOS. This may be attributed to the limited search space in small instances, allowing algorithms to converge on comparable near-optimal solutions. Since the SOS is search-based, it has a slight advantage in limited search spaces. We will further investigate this hypothesis in the next chapter by evaluating our approach in larger instances. Secondly, our approach offers a clear advantage over SOS in terms of computation time. Since PetriRL is a learning-based algorithm, it develops a strategy (or policy) during training. Once trained, the RL agent can apply the policy directly without performing an exhaustive search, as metaheuristic methods like SOS require. As a result, PetriRL requires only a fraction of the time to reach a solution. In the tested instances, the average computation time for PetriRL was 0.24 seconds, compared to 217 seconds for SOS.

\begin{figure}[ht]
\centering
\includegraphics[width=\linewidth]{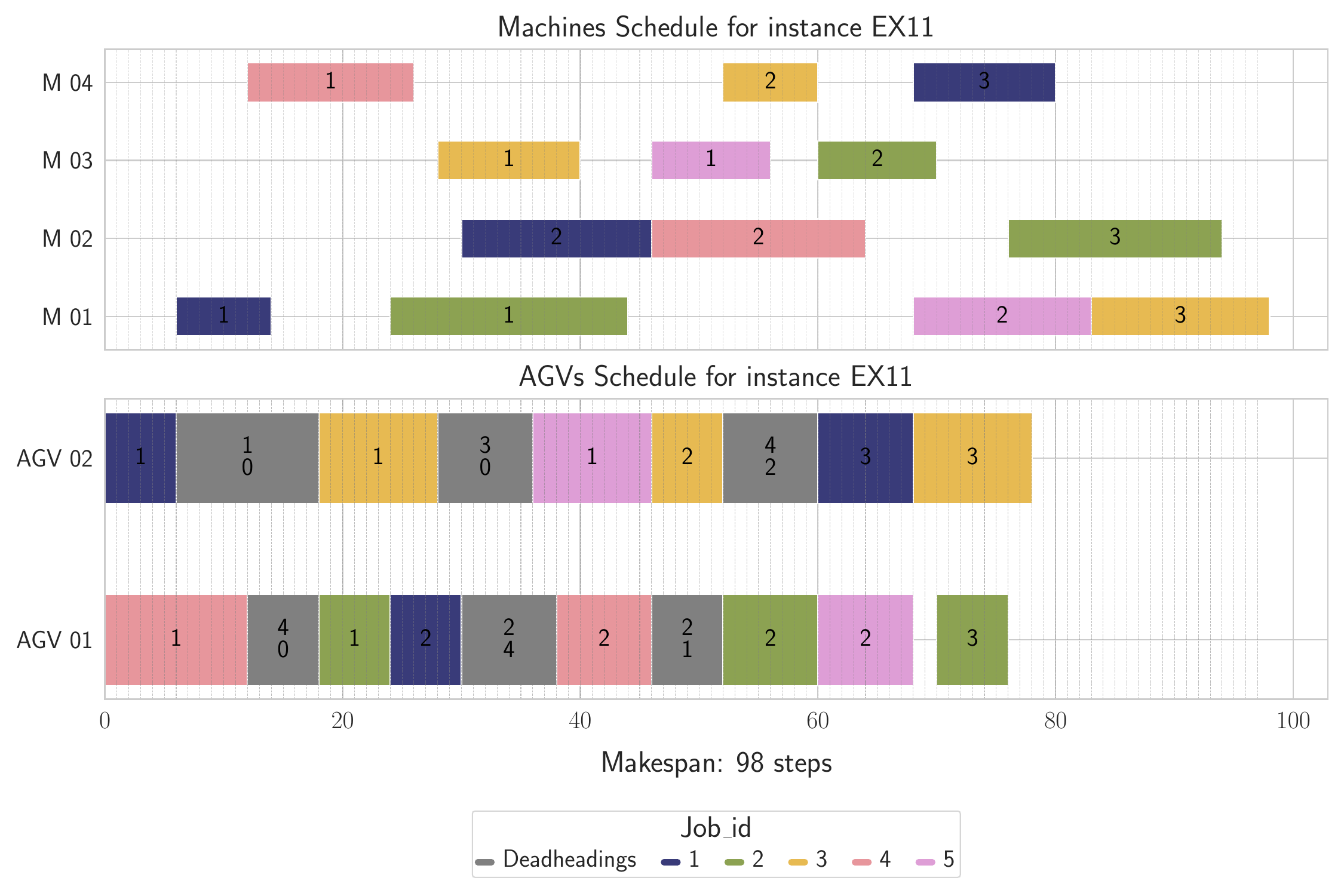}
\caption{Gantt chart solution for the instance "EX11" (5 jobs × 4 machines). The top chart shows machine allocation schedules, and the bottom chart shows AGV allocation schedules. Colors represent different jobs, with numbers indicating the operation sequence within each job. Grey segments in the AGVs schedule denote deadheading, with the origin at the top and the destination at the bottom.}
\label{fig: Gannt small}
\end{figure}

In Fig~\ref{fig: Gannt small}, we present an example of a Gantt chart schedule generated by our algorithm for a small instance, "EX11" (5 jobs × 4 machines). Each color represents a distinct job, with numbered annotations indicating the operation token order in the job. In the AGVs schedule, grey segments represent deadheading, where an AGV moves between machines without transporting a part. Unlike traditional JSSP, optimization in this context heavily depends on AGV allocation, which is influenced by both the sequence of operations and the AGV assignment. Minimizing deadheading is essential for reducing makespan, as effective scheduling can significantly reduce unnecessary AGVs' travel. For instance, if an AGV can deliver a part to a machine and immediately pick up another part from the same machine, it eliminates deadheading. This can be observed at step 21, where AGV 01 efficiently services jobs 1 and 2 consecutively without needing to relocate, optimizing its operational efficiency.

\subsubsection{AGVs and tool sharing implementation on a large instance}

After conducting an initial round of testing on small instances, we increased the complexity, testing on the newly proposed large instances introduced in Section \ref{section: new benchmark}, and incorporating tool sharing. In the previous section, the results showed that SOS was the best-performing approach in small instances. Since the algorithm was only tested on small instances in the original publication, we first replicate the SOS algorithm to validate its fidelity to the original implementation, as detailed in Section \ref{section: SOS replication}. Once confirmed, we apply the replicated code to the large instances as our primary benchmark.

\begin{table*}[ht]
  \centering
  \resizebox{\linewidth}{!}{%
  \begin{tabular}{ccccccccccccc} 
    \toprule
    & \multicolumn{2}{c}{} 
    & \multicolumn{5}{c}{\textbf{AGVs only} }
    & \multicolumn{4}{c}{\textbf{AGVs and tools sharing}} \\
    
    \cmidrule(lr){3-7} \cmidrule(lr){8-13}

    Instance & Size JxM)   &  PetriRL &  Time (s)  & Meta SOS  &  Time (s) &  Gap & &   PetriRL  &  Time (s)   & Meta SOS  &  Time (s) & Gap \\
    \midrule
     sl00     & 15x15      & 1801      & 5.50       & 2194      & 600       & 18\%      & &    2639     & 14.48       &   4057    &   600      & 35\%  \\
     sl10     & 20x15      & 2057      & 7.44       & 2483      & 600       & 17\%      & &    3163     & 17.29       &   5146  &   600   & 39\%  \\
     sl20     & 20x20      & 2619      & 11.54       & 3448      & 600       & 24\%      & &    4148     & 26.26       &   6179  &   600   & 33\%     \\   
     sl30     & 30x15      & 2436      & 11.38       & 3250      & 600       & 25\%      & &    5499     & 28.94       &   7635  &   600   & 40\%  \\          
     sl40     & 30x20      & 3178      & 17.64      & 4194      & 600       & 24\%      & &    5989     & 37.61       &   9065  &   600   & 34\%     \\          
     sl50     & 50x15      & 3478      & 27.64     & 4703      & 600       & 26\%      & &    7122     & 48.50       &   12558 &   600   & 43\%     \\
     sl60     & 50x20      & 4139      & 37.65      & 5837      & 600       & 29\%      & &   8755     & 72.27       &   15765 &   600   & 44\%     \\
     sl70     & 100x20     & 7623      & 64.09      & 9738      & 600       & 22\%      & &   18845    & 144.12      &   30537 &   600   & 38\%     \\

    \bottomrule
  \end{tabular}
  }
  \caption{Comparison of Makespan and Computation Time between PetriRL and SOS on a large instance, evaluating both AGVs-only and AGVs + Tool-sharing implementations.}
  \label{table: Results_sos}
\end{table*}

The results in Table \ref{table: Results_sos} present the makespan for PetriRL and SOS and their respective computation times. Since SOS is search-based, its computation time heavily depends on the instance size. For example, solving a (5 jobs × 4 machines) instance took 128 seconds, while solving a (15 jobs × 15 machines) instance took 31 hours. To address this, we decided to limit the SOS search time to 10 minutes (600 seconds) on large instances. The results highlight the advantage of our approach on large instances. PetriRL outperforms SOS, with improvements ranging from 18\%  to 25\%  in the AGVs-only implementation and from 35\% to 40\% in the AGVs and tools-sharing implementation.

\begin{figure}[ht]
\centering
\includegraphics[width=\linewidth]{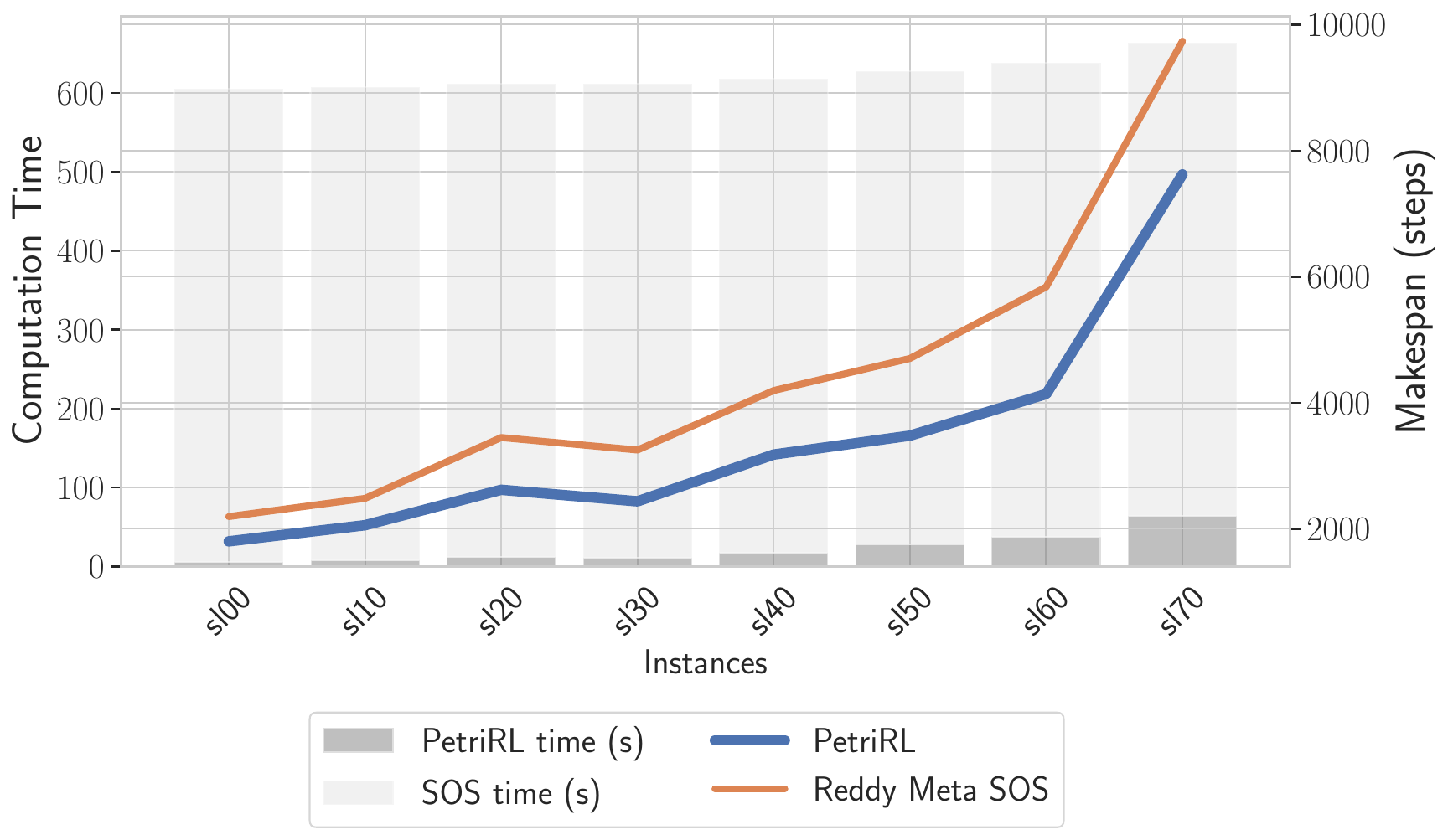}
\caption{Performance comparison across various instances using SOS metaheuristics vs. our approach, PetriRL. The line plots show makespan results for each method, while the bar plot compares the computation time between our method and SOS.}
\label{fig: large_instances}
\end{figure}

The AGVs-only results in large instances plotted in Figure~\ref{fig: large_instances} show that our approach outperforms the SOS metaheuristic using a fraction of the computation time. This can be explained by the fact that purely search-based approaches, such as SOS, lose their competitive edge in larger instances. As the number of machines and jobs increases, the search space expands exponentially, making it unfeasible to explore the entirety of the search space within polynomial time. This demonstrates the NP-hard nature of the FMS  problem. On the other hand, learning-based approaches like PetriRL only require a search during the training phase. Once a policy is trained, the algorithm can leverage the acquired knowledge to make decisions without re-exploring the search space. This characteristic is crucial in this case and for adapting to frequent changes without conducting a full search. This adaptability is vital for solving dynamic scheduling problems like FMS, which will be further highlighted in the subsequent chapters.

\begin{figure}[ht]
\centering
\includegraphics[width=\linewidth]{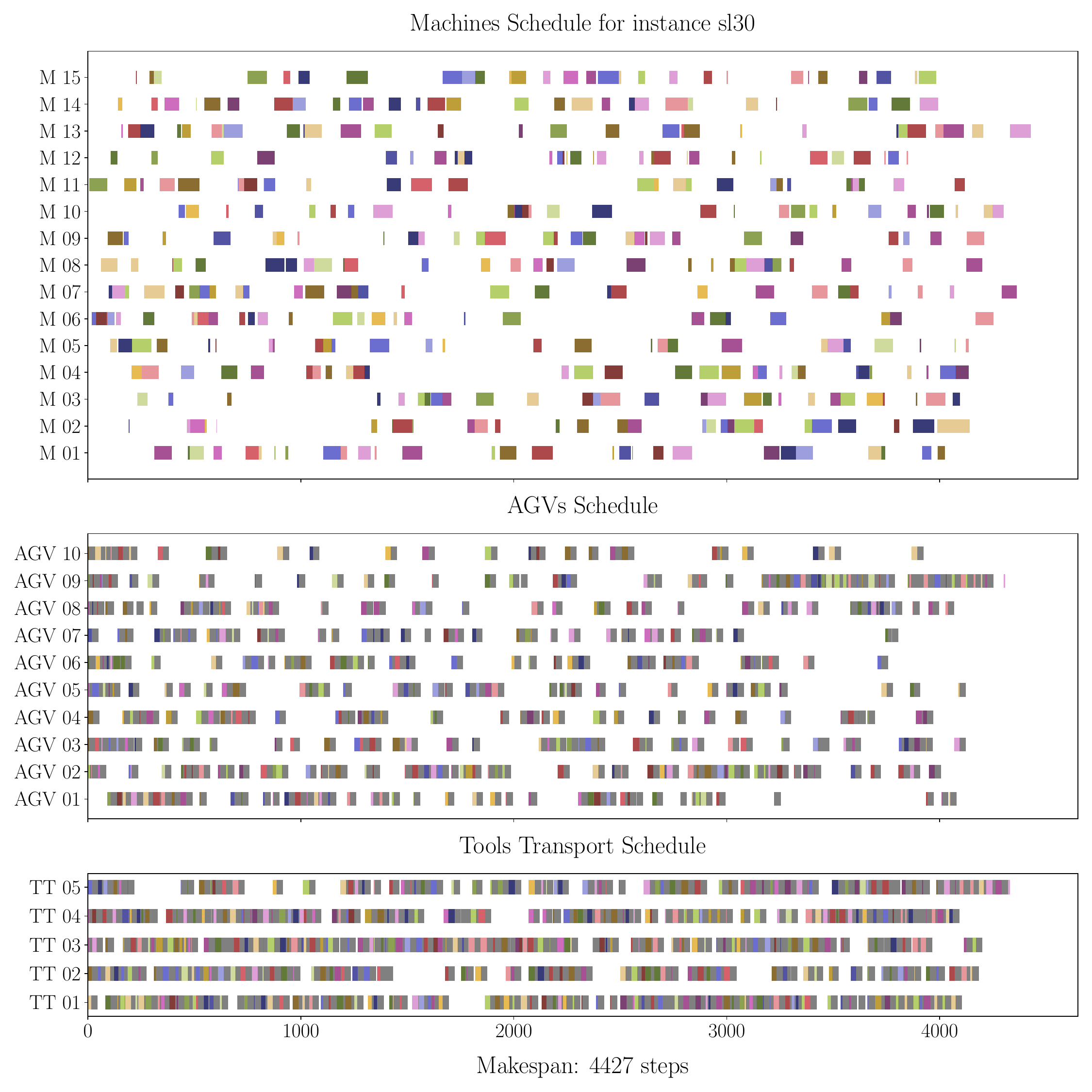}
\caption{Gantt chart solution for the instance "sl30" (30 jobs × 15 machines). The top chart shows machine allocation schedules, the middle chart shows the AGVs allocation schedule, and the bottom chart shows the schedule of the tool transporters.}
\label{fig: Gannt large}
\end{figure}

As with the small instances, we present in Fig~\ref{fig: Gannt large} an example of a Gantt chart schedule generated by our algorithm for a larger instance, "sl30" (30 jobs × 15 machines × 15 tools), using 10 AGVs and five tool transporters. From the graph, we observe that the tool transporters are the bottleneck in the system. This is evident as a machine cannot begin processing an operation until both the piece to be processed is delivered by the AGV and the required tool is delivered by the tool transporter. The interaction between AGVs and tool transporters introduces a significant constraint, emphasizing the importance of effective scheduling and resource allocation to optimize the system's performance.

\subsection{Generalization performance}
\label{section: Generalization}

In this section, we evaluate the generalization performance of our approach by comparing two strategies: training a single agent capable of handling all variable sizes versus employing a dedicated agent specifically trained for each instance size group. 

Leveraging the flexible structure of the Petri Net, our approach can efficiently solve a 15-job x 15-machine problem using a 100-job x 20-machine Petri Net layout. Starting with a 100-job x 20-machine empty Petri Net layout, the algorithm sequentially fills job slots with the tokens from the 15-job x 15 instance, leaving the remaining slots empty. Empty job operation slots disable their corresponding job selection transitions, which results in them being automatically masked, reducing the agent's action space. Likewise, unused machine buffer slots are also automatically masked. The observation is padded with null values for unoccupied slots, allowing the agent to handle varying instance sizes and adapt to configurations without sacrificing performance. Since the employed algorithm is policy-based, the agent can seamlessly manage variable sizes, interpreting a fully populated 15x15 instance as a mid-completed 100x20 instance.

\begin{figure}[H]
\centering
\includegraphics[width=\linewidth]{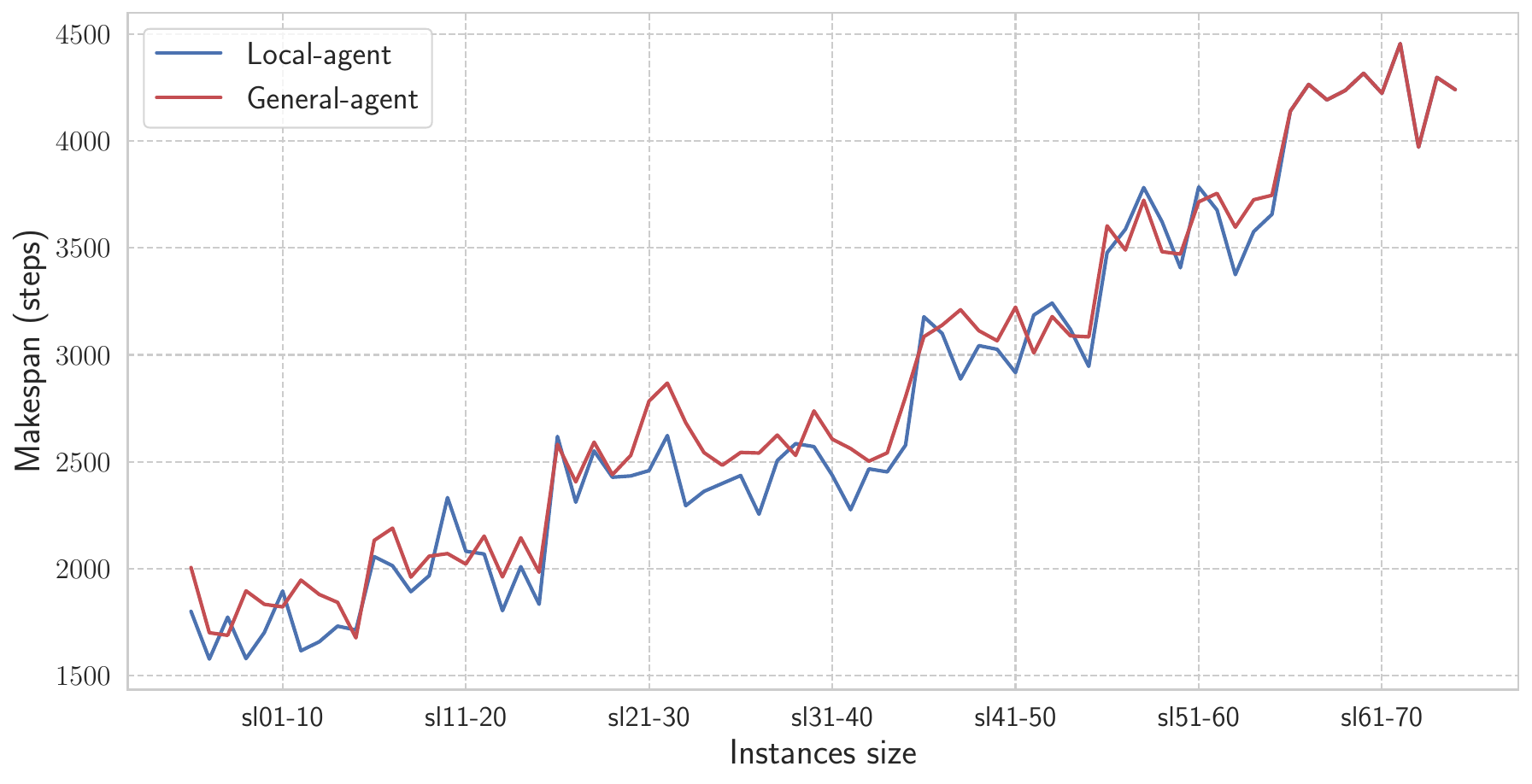}
\caption{Generalizability performance across different instances using general and specialized agents. The general agent (in red) is trained solely on the "sl60" (50 × 20) instance and tested on 60 other benchmark instances. For the specialized agents (in blue), six agents are each trained on a specific size group and tested on ten instances within the same size category.}
\label{fig: generalization}
\end{figure}

The results in Fig~\ref{fig: generalization} demonstrate strong generalization performance across seven distinct size groups, covering 70 instances. While the general agent exhibited a slight decrease in performance with a 4\% increase in makespan average compared to specialized agents, it achieved this performance without requiring size-specific agent training. This highlights a trade-off between generalizability and optimal performance.

\subsection{Dynamic Behavior}

One of the key distinctions between SOS and other metaheuristics, compared to the policy-based RL approach used in this paper, is their problem-solving approach. Metaheuristics are predominantly guided search-based, whereas RL approaches are learning-based. Although search-based metaheuristics excel in thoroughly exploring the search space, they lack inherent knowledge retention, requiring a fresh search with every change in the instance. In contrast, the RL framework balances search during its exploration phase with knowledge retention through its value and policy networks, especially in actor-critic algorithms. This is particularly advantageous in dynamic environments like FMS, where an RL agent can adapt to changes by following its learned policy without the need to be completely retrained. 

\begin{figure}[H]
\centering
\includegraphics[width=\linewidth]{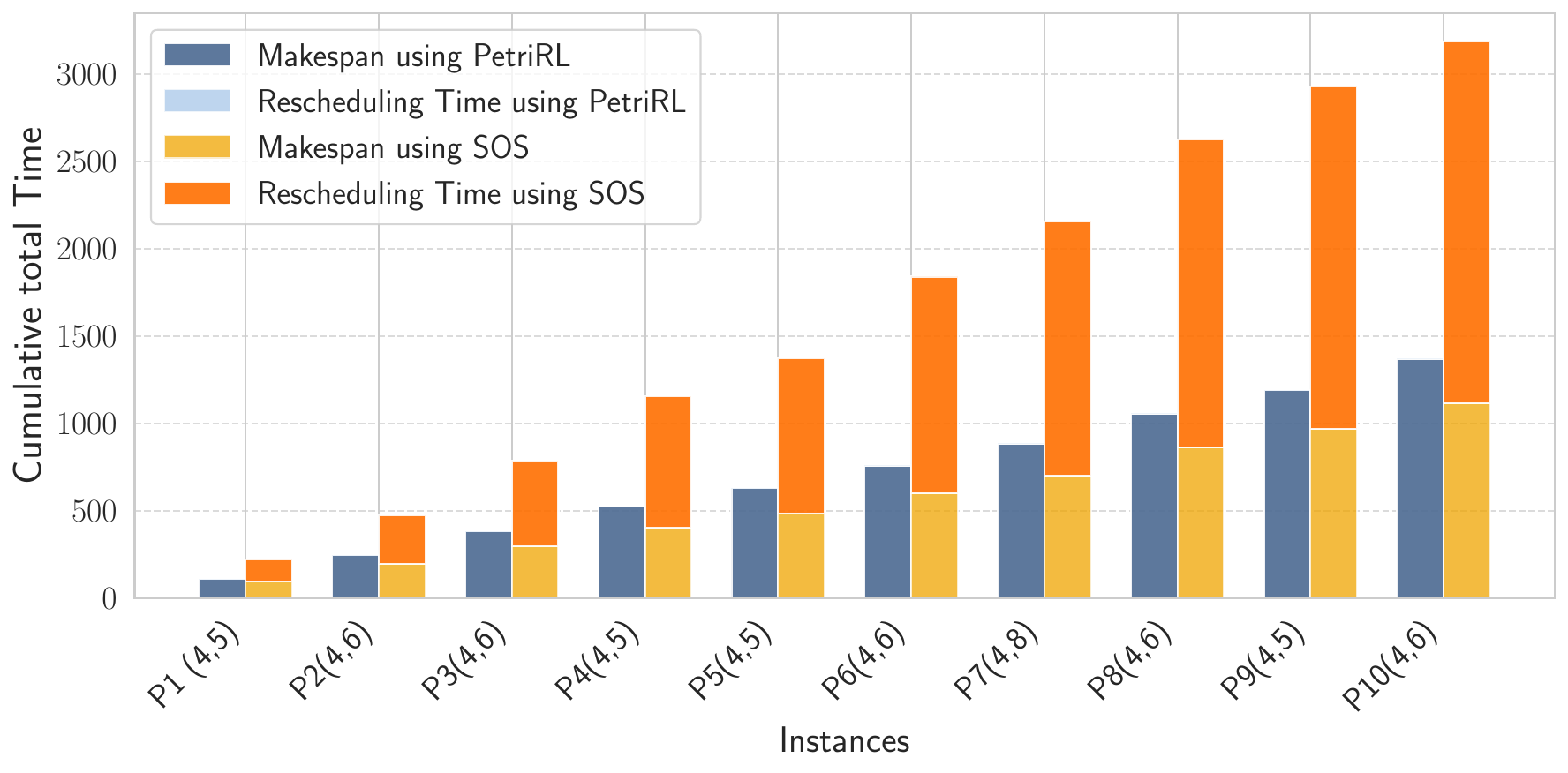}
\caption{Comparison of the cumulative calculation time between SOS and PetriRL in sequential instance injection. The bar plots show the cumulative makespan and computation time required for all preceding instances.}
\label{fig: dynamic}
\end{figure}

To illustrate this observation, we propose a scenario where instances are divided into batches and sequentially fed to the SOS metaheuristic optimizer and our RL-based optimizer. This sequential batch processing simulates scenarios such as late job arrivals or changes in job order. We consider the scenario where ten partitions from P1 to P10 are fed sequentially to the systems, and we log the total time, including the computation time. For the RL-based optimization, we used the global agent approach introduced in section~\ref{section: Generalization}, where one agent is trained to handle different-sized instances, including the different-sized partitions in this case. The results are shown in Fig~\ref{fig: dynamic}.

When rescheduling time is at a first stage disregarded, and only makespan is considered, the SOS approach initially outperforms the RL-based PetriRL optimizer due to its exhaustive search capabilities within constrained search spaces, often achieving a more optimal solution. This is particularly evident in the final partition, P10, as shown in Fig~\ref{fig: dynamic}. However, when rescheduling time is considered, the limitations of SOS become clear; it necessitates a full restart of its search for each new sub-instance. In contrast, leveraging its policy network, the RL agent adapts dynamically to changes in job sequencing without retraining, seamlessly adjusting to evolving conditions. This adaptability enables the RL agent to achieve superior combined search and processing times, particularly as instances become more fragmented. These results suggest that learning-based approaches are inherently better suited to solve dynamic frameworks like FMS, where rapid adjustment and flexibility are essential, compared to more static ones.

\subsection{Ablation Study}

In this section, we conduct an ablation study to understand the contribution of individual components to the results. We highlight three components whose combination contributes to our approach performance: reward shaping, lookahead, and invalid actions masking.

\begin{table}[ht]
  \centering
    \resizebox{\linewidth}{!}{%
    \begin{tabular}{ccc|ccc}
      \toprule
      \multicolumn{3}{c}{\textbf{Effect of Reward Shaping }} & \multicolumn{3}{c}{\textbf{Effect of Lookahead}} \\
      \cmidrule(lr){1-3} \cmidrule(lr){4-6}
      \textbf{Instance} & \textbf{With } & \textbf{Without} & \textbf{Instance} & \textbf{With} & \textbf{Without } \\
      \midrule
      sl40 & 3178 & 3278 & EX12 & 82 & 100 \\
      sl41 & 3100 & 3435 & EX22 & 84 & 94 \\
      sl42 & 2887 & 3160 & EX32 & 91 & 99 \\
      sl43 & 3043 & 3113 & EX42 & 95 & 108 \\
      sl44 & 3026 & 3094 & EX52 & 72 & 73 \\
      sl45 & 2917 & 3130 & EX62 & 104 & 124 \\
      sl46 & 3186 & 3329 & EX72 & 92 & 105 \\
      sl47 & 3242 & 3499 & EX82 & 151 & 151 \\
      sl48 & 3119 & 3259 & EX92 & 105 & 110 \\
      sl49 & 2946 & 3085 & EX102 & 143 & 153 \\
      \midrule
      \textbf{Average} & \textbf{3064.4} & \textbf{3238.2} &  \textbf{Average}& \textbf{102} & \textbf{112} \\
      \bottomrule
    \end{tabular}%
    }

  \caption{The effect of reward shaping on midsized instances (30 jobs x 20 machines) and the effect of lookahead on small instances on the makespan.}
  \label{table: combined_effects}
\end{table}

Starting with the effect of reward shaping, we compare two reward functions: machine utilization as a form of reward shaping and makespan as a sparse reward. As explained in Section \ref{section: environment}, the sparse reward problem is more pronounced in large instances where the action sequence leading to termination is longer, exacerbating the credit assignment problem. In Table \ref{table: combined_effects}, we tested makespan and machine utilization as reward functions on mid-sized instances (30 jobs, 20 machines). The results show that reward shaping improves credit assignment, leading to a better policy and a lower average makespan. The contribution of lookahead, introduced in Section \ref{section: lookahead}, was tested on small-sized instances. The results indicate that lookahead improves AGVs' positioning, which in turn improves overall performance.


\begin{figure}[ht]
\centering
\includegraphics[width=\linewidth]{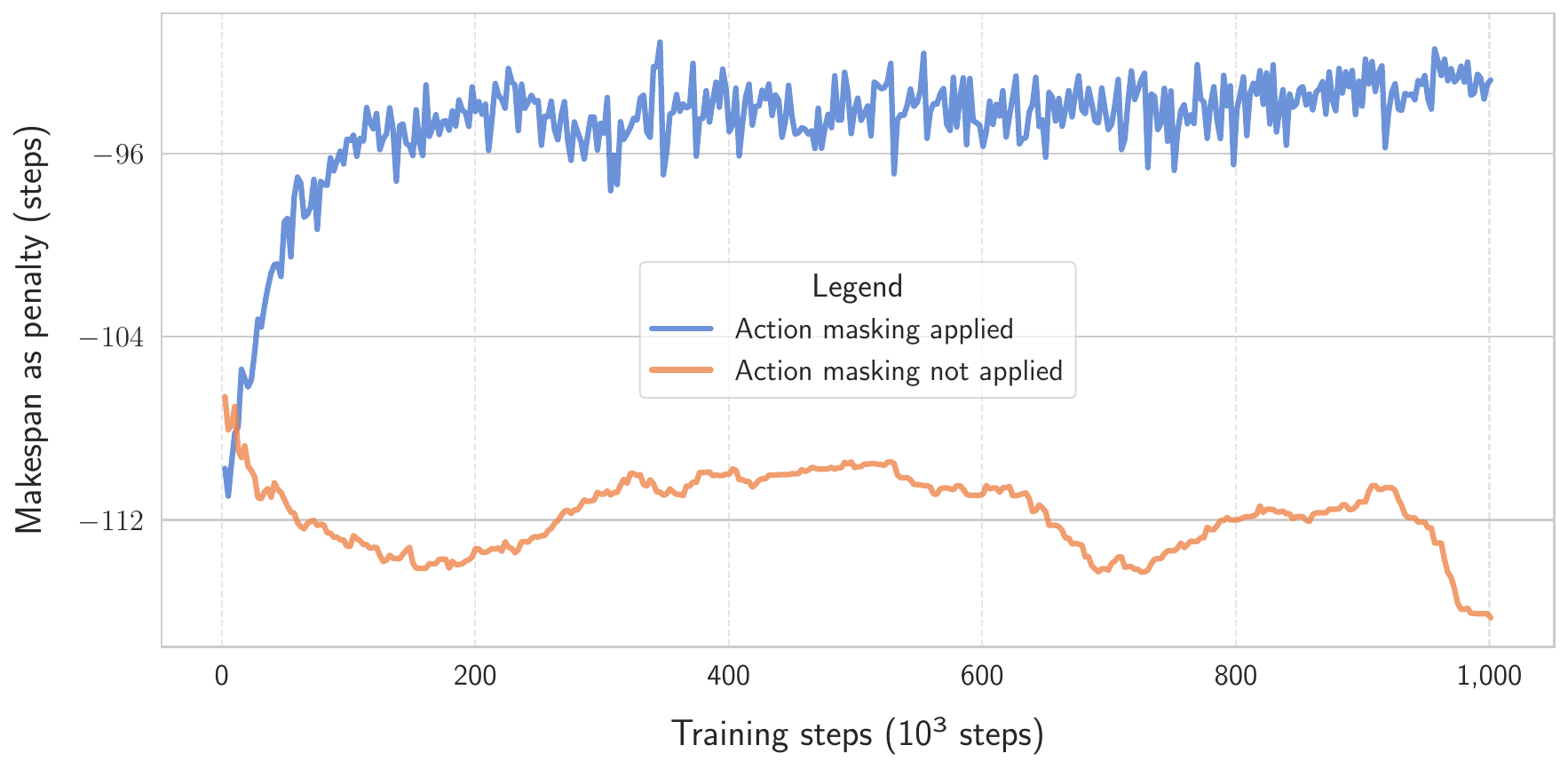}
\caption{Ablation study: Effect of action masking on training performance. The blue line represents the reward evolution during training with action masking enabled, while the orange line shows the reward evolution when action masking is not applied.}

\label{fig: ablation}
\end{figure}

In Figure~\ref{fig: ablation}, we combined dynamic masking via the Petri Net guard function with maskable PPO and compared training performance with and without masking to assess the impact of invalid actions masking. Action masking significantly improves training by reducing the search space and accelerating learning. It also enhances credit assignment by excluding irrelevant actions, aiding in efficient policy optimization. The results show that the reward steadily increases and stabilizes after 200,000 steps with masking. In contrast, without masking, the reward remains unstable and fails to reach the same level, highlighting the challenges in training without masking.

m\section{Conclusion }
\label{section: Conclusion}

In this paper, we propose a novel approach to addressing FMS by integrating the formal modeling strengths of Petri Nets with the dynamic decision-making capabilities of RL. This hybrid solution offers a modular and interpretable framework, allowing for flexible integration and testing of various system components such as AGVs, shared tools, and machines. Motivated by a gap in the literature around large, realistic instances that include scenarios with AGVs and tool-sharing, we introduced a new benchmark inspired by the Taillard benchmark, incorporating both tool-sharing and transportation complexities for large instances.

Our approach was evaluated incrementally, starting with smaller instances. It outperformed traditional heuristics and achieved competitive results compared to metaheuristics, with significantly reduced computation times. Unlike metaheuristic methods, our RL-based approach benefits from knowledge retention, eliminating the need for an exhaustive search after training. Further tests with complex scenarios involving AGVs and tool transport demonstrated that our method surpasses search-based heuristics on large instances, achieving solutions in a fraction of the time. We also examined our approach's adaptability and generalizability to validate its robustness. Results indicated that our model is better equipped to handle dynamic changes in the manufacturing environment than metaheuristics, establishing it as a more effective solution for FMS. Additionally, our approach demonstrated strong generalization capability, allowing it to manage a range of instance sizes with minimal performance degradation, thus providing a scalable framework. An ablation study revealed the importance of several components in improving performance. Reward shaping significantly improved credit assignment, leading to a more effective policy and a lower average Makespan. Lookahead enhanced AGVs positioning, boosting overall performance in small-sized instances. Action masking is achieved by dynamically filtering out irrelevant actions, accelerating learning, and optimizing credit assignments. 

This study demonstrates the adaptability of our approach in responding to external dynamic changes, like varying incoming orders, laying a solid foundation for future research. It opens avenues to explore how the model can adjust to internal environmental changes, including unforeseen equipment breakdowns, scheduled maintenance, and strategic tool changes, particularly when machining advanced materials with high strength. The model can account for tool replacements based on wear patterns, enhancing system robustness and operational efficiency in real-world manufacturing environments.


\bibliographystyle{elsarticle-num} 
\bibliography{References.bib}


\end{document}